\documentclass{article}
\usepackage{arxiv}
\usepackage{hyperref}       
\usepackage{url}
\usepackage{booktabs}       
\usepackage{amsfonts}
\usepackage{nicefrac}
\usepackage{microtype}   
\usepackage{lipsum}
\usepackage{graphicx}
\usepackage{verbatim}
\usepackage{multirow}
\usepackage{multicol}
\usepackage{color}
\usepackage{latexsym}
\usepackage{graphicx}
\usepackage{xcolor}
\usepackage{colortbl,booktabs}
\usepackage{tabu}
\usepackage{float}
\usepackage{amsmath}
\usepackage{wrapfig}
\usepackage{multirow}

\usepackage{CJKutf8}
\usepackage[utf8]{inputenc} % allow utf-8 input
\usepackage[T1]{fontenc} % use 8-bit T1 fonts
\usepackage{CJK}
\makeatletter
\def\UrlAlphabet{%
      \do\a\do\b\do\c\do\d\do\e\do\f\do\g\do\h\do\i\do\j%
      \do\k\do\l\do\m\do\n\do\o\do\p\do\q\do\r\do\s\do\t%
      \do\u\do\v\do\w\do\x\do\y\do\z\do\A\do\B\do\C\do\D%
      \do\E\do\F\do\G\do\H\do\I\do\J\do\K\do\L\do\M\do\N%
      \do\O\do\P\do\Q\do\R\do\S\do\T\do\U\do\V\do\W\do\X%
      \do\Y\do\Z}
\def\UrlDigits{\do\1\do\2\do\3\do\4\do\5\do\6\do\7\do\8\do\9\do\0}
\g@addto@macro{\UrlBreaks}{\UrlOrds}
\g@addto@macro{\UrlBreaks}{\UrlAlphabet}
\g@addto@macro{\UrlBreaks}{\UrlDigits}
\makeatother

\title{\textsc{ERNIE 3.0}: Large-scale Knowledge Enhanced Pre-training for Language Understanding and Generation}

\author{Yu Sun\thanks{Equal Contribution} \and \textbf{Shuohuan Wang}$^{*}$ \and \textbf{Shikun Feng}$^{*}$ \and \textbf{Siyu Ding} \and \textbf{Chao Pang}\\
\and \textbf{Junyuan Shang~~~~Jiaxiang Liu~~~~Xuyi Chen~~~~~Yanbin Zhao~~~~~Yuxiang Lu~~~~~Weixin Liu}  \\ 
\and \textbf{Zhihua Wu~~~~~~~~Weibao Gong~~~~~~~~~Jianzhong Liang~~~~~~~~Zhizhou Shang~~~~~~~~Peng Sun} \\ \
\and \textbf{Wei Liu~~~~~~~Xuan Ouyang~~~~~~~Dianhai Yu~~~~~~~Hao Tian~~~~~~~Hua Wu~~~~~~~Haifeng Wang}\\  \\

\bf{ Baidu Inc}. \\ \\
 \{\texttt{sunyu02, wangshuohuan, fengshikun01}\}\texttt{@baidu.com}
  \\
}

\begin{document}
\maketitle
\begin{abstract}
Pre-trained models have achieved state-of-the-art results in various Natural Language Processing (NLP) tasks. 
Recent works such as T5~\cite{T5} and GPT-3~\cite{gpt-3} have shown that scaling up pre-trained language models can improve their generalization abilities. 
Particularly, the GPT-3 model with 175 billion parameters shows its strong task-agnostic zero-shot/few-shot learning capabilities. 
Despite their success, these large-scale models are trained on plain texts without introducing knowledge such as linguistic knowledge and world knowledge. 
In addition, most large-scale models are trained in an auto-regressive way. As a result, this kind of traditional fine-tuning approach demonstrates relatively weak performance when solving downstream language understanding tasks. 
In order to solve the above problems, we propose a unified framework named ERNIE 3.0 for pre-training large-scale knowledge enhanced models. It fuses auto-regressive network and auto-encoding network, so that the trained model can be easily tailored for both natural language understanding and generation tasks with zero-shot learning, few-shot learning or fine-tuning. 
We trained the model with 10 billion parameters on a 4TB corpus consisting of plain texts and a large-scale knowledge graph.
Empirical results show that the model outperforms the state-of-the-art models on 54 Chinese NLP tasks, and its English version achieves the first place on the SuperGLUE~\cite{wang2019superglue} benchmark (July 3, 2021), surpassing the human performance by +0.8\% (90.6\% vs. 89.8\%).

\end{abstract}
\section{Introduction}
Pre-trained language models such as ELMo~\cite{peters2018deep}, GPT~\cite{radford2018improving}, BERT~\cite{devlin2018bert}, and ERNIE~\cite{sun2019ernie} have proved to be effective for improving the performances of various natural language processing tasks including sentiment classification \cite{socher2013recursive}, natural language inference \cite{bowman2015large}, text summarization \cite{hu2015lcsts}, named entity recognition \cite{sang2003introduction} and so on. In general, pre-trained language models are learned on a large amount of text data in a self-supervised manner, and then fine-turned on downstream tasks or directly deployed through zero/few-shot learning without task-specific fine-tuning. Such pre-trained language models have become the new paradigm for natural language processing tasks.

% introduce the trend of large scale pre-trained model
In the past year or two, one of the important trends of pre-trained language models is their increasing model size, which leads to lower perplexity in pre-training and better performances on downstream tasks. Megatron-LM \cite{Megatron-LM}, with one billion parameters, is proposed for language understanding using a simple but efficient intra-layer model parallel approach, which achieves the state-of-the-art results on several datasets. T5 \cite{T5} explores the limits of pre-trained models with 10 billion parameters, but soon the record was broken by the GPT-3 model~\cite{gpt-3} with 175 billion parameters which has a good performance under the few-shot or even zero-shot settings. Soon afterwards, Switch-Transformer~\cite{switch-transformer} is proposed as the world's first trillion-parameter pre-trained language model.

However, these large-scale pre-trained language models with hundreds of billions of parameters are trained on plain texts. For example, the 175-billion-parameter GPT-3 is trained on a corpus with 570GB filtered texts from Common Crawl. Such raw texts lack explicit representation of knowledge such as linguistic knowledge and world knowledge. In addition, most large-scale models are trained in an auto-regressive way, but \cite{devlin2018bert} shows that such models demonstrate poorer performance with traditional fine-tuning when adapting to downstream language understanding tasks.

In this work, to solve the problem caused by a single auto-regressive framework and to explore the performance of knowledge enhanced pre-trained models with large-scale parameters, we propose a unified framework called ERNIE 3.0 to train large-scale knowledge enhanced models on a 4TB corpus consisting of plain texts and a large-scale knowledge graph by fusing the auto-regressive network and the auto-encoding network. The proposed ERNIE 3.0 can handle both natural language understanding tasks and natural language generation tasks through zero-shot learning, few-shot learning or fine-tuning. Furthermore, the proposed framework supports the introduction of various customized tasks at any time. These tasks share the same encoding networks and are trained through multi-task learning. This method makes the encoding of lexical, syntactic and semantic information across different tasks possible. Moreover, when given a new task, our framework could incrementally train the distributed representations based on the previous training parameters, with no need to train them from scratch.

In summary, our contributions are as follows:
\begin{itemize}
    \item We propose a unified framework ERNIE 3.0, which combines auto-regressive network and auto-encoding network so that the trained model can handle both natural language understanding and generation tasks through zero-shot learning, few-shot learning or fine-tuning. 
    \item We pre-train large-scale knowledge enhanced models with 10 billion parameters and evaluate them with a series of experiments on both natural language understanding and natural language generation tasks. Experimental results show that ERNIE 3.0 consistently outperforms the state-of-the art models on 54 benchmarks by a large margin and achieves the first place on the SuperGLUE~\cite{wang2019superglue} benchmark. 
 
\end{itemize}

\section{Related Work}
\subsection{Large-scale Pre-trained Models}

Since BERT \cite{devlin2018bert} is proposed as a powerful language model for natural language understanding, pre-trained language models have attracted more and more attention and become the new paradigm for natural language processing. One of the research trends is increasing model size, which leads to lower perplexity and better performance~\cite{kaplan2020scaling}. As a result, many large-scale pre-trained models have been proposed in the past two years. T5 model \cite{T5} is proposed to push the performance for both natural language understanding and natural language generation tasks with 11 billion parameters. The T5 model converts all text-based language tasks into a text-to-text format by a unified framework and fully explores the effectiveness of pre-training objectives, architectures, unlabeled datasets, transfer approaches, and other factors. After the T5 model, GPT-3 \cite{gpt-3}, which includes 175 billion parameters, is proposed to achieve an amazing performance on a wide range of tasks under the few-shot and zero-shot settings. Specifically, GPT-3 is an auto-regressive language model, 10x more than its predecessor, GPT-2, proposed by \cite{radford2019language}. However, GPT-3 shows a lack of common sense, exists biases and privacy issues in the tests \cite{rise_risk_gpt3}. \cite{switch-transformer} have proposed a 1 trillion parameters model named Switch Transformer with simplifying MoE~\cite{jacobs1991adaptive, jordan1994hierarchical} routing algorithm to improve model with less communication and computational costs, and \cite{switch-transformer} also proposed a large scale distributed training solution to tackle the problem of training complexity, communication costs, and training instability.

Besides the models mentioned above, more non-English large models have been proposed recently. \cite{zhang2020cpm} released a 2.6 billion parameters Chinese Pre-trained Language Model (CPM) with generative pre-training on large-scale Chinese training data and the model structure was inspired by \cite{gpt-3}.  \cite{zhang2021cpm} have released a 11 billion parameters model CPM-2. To accelerate the pre-training based on existing PLMs instead of training models from scratch, the knowledge inheritance techniques have been introduced and during the fine-tuning stage, prompt tuning is involved to better exploit the knowledge within the pre-trained model. \cite{lin2021m6} have proposed a cross-modal pre-training method called M6({M}ulti-{M}odality to {M}ulti-{M}odality {M}ultitask {M}ega-Transformer) including 100 billion parameters for unified pre-training on multiple modalities data. \cite{zeng2021pangu} proposed a 200 billion parameters auto regressive language model named PangGu-$\alpha$ which is trained on a cluster of 2048 Ascend 910 AI processors with distributed training techniques including data parallelism, op-level model parallelism, pipeline model parallelism, optimizer model parallelism and re-materialization. Except for those Chinese large-scale models, a Korean 204 billion parameters language model named HyperCLOVA  \cite{HyperClova} has been proposed, and its volume of machine-learned data in Korean was 6,500 times larger than GPT-3’s. From what has been discussed above, observations now suggest that large-scale pre-trained models have attracted more and more attention from industry and academia.

\subsection{Knowledge Enhanced Models}
Pre-trained language models capture syntactical and semantic knowledge from large-scale corpus, but lack world knowledge. Recently, several works have attempted to incorporate world knowledge in pre-trained language models. The typical form of world knowledge is a knowledge graph. Many works (\cite{zhang2019ernie, peters2019knowledge, he2019integrating}) integrate entity and relation embedding from knowledge graph in pre-trained language models. WKLM~\cite{xiong2019pretrained} replaced entity mentions in the original documents with names of other entities of the same type and train the models to distinguish the correct entity mention from randomly chosen ones. KEPLER~\cite{wang2021kepler} optimized the models with knowledge embedding and mask language model objectives to align the world knowledge and language representation into the same semantic space. CoLAKE~\cite{sun2020colake} integrated the language context and the knowledge context in a word-knowledge graph and jointly learned contextualized representation for language and knowledge with the extended mask language model objective. Another existing form of world knowledge is the extra annotation of large-scale data. ERNIE 1.0~\cite{sun2019ernie} introduced phrase masking and named entity masking and predicts the whole masked phrases and named entities to help the model learn the dependency information in both local contexts and global contexts. CALM~\cite{zhou2020pre} teached models to detect and revised a corrupted sentence with the incorrect ordering of concepts and to distinguish truth sentences from less plausible ones via two kinds of self-supervised pre-training tasks. K-Adapter\cite{wang2020k} utilized adapters trained on different knowledge sources with extra annotations to distinguish where the knowledge comes from.

\section{ERNIE 3.0}

 \begin{figure*}[ht]
	\centering
	\includegraphics[width=1.0\textwidth]{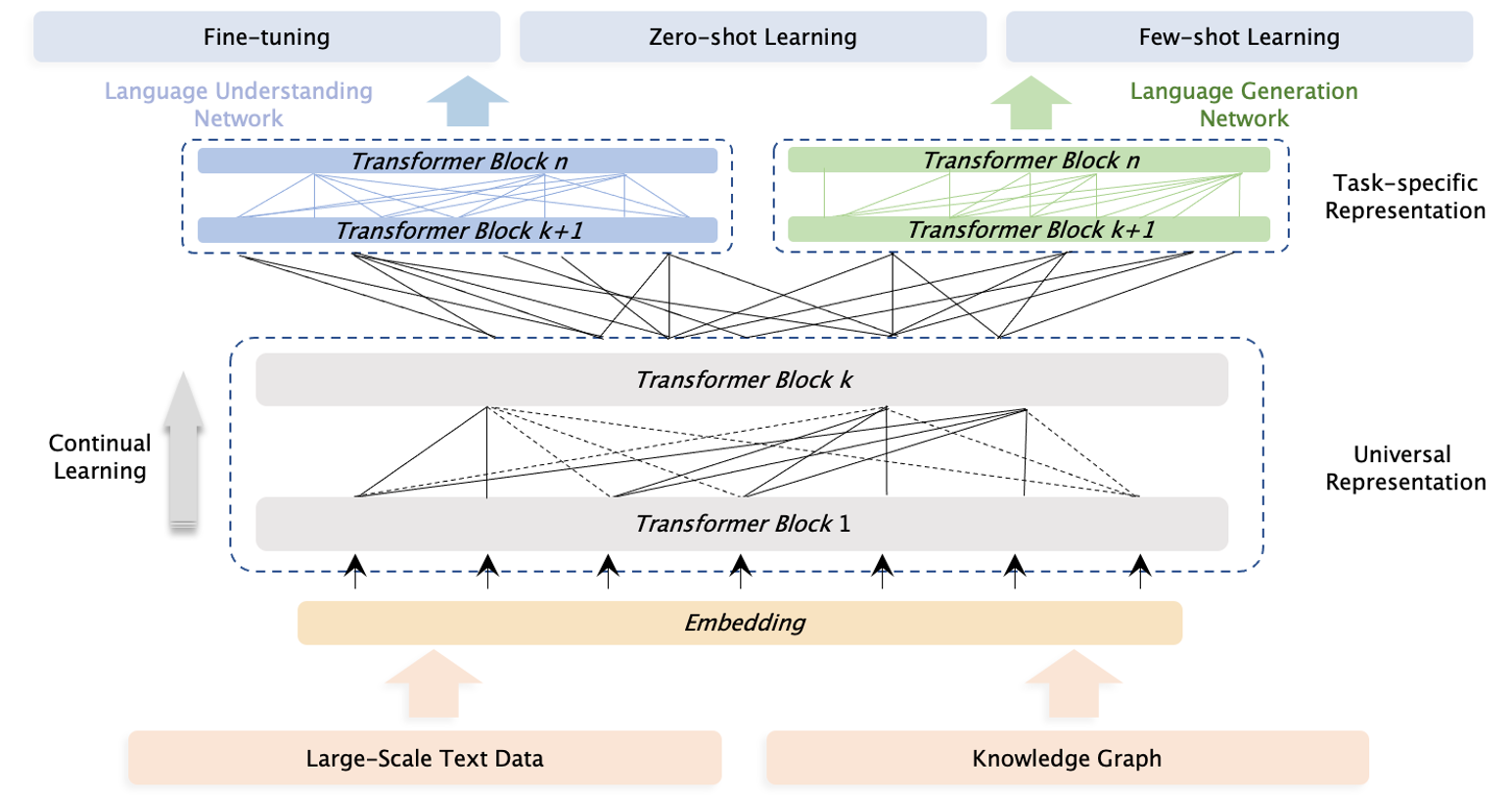}
	\caption{The framework of ERNIE 3.0.} \label{label:ernie 3.0 framework}
\end{figure*}

A significant improvement has been achieved on various natural language processing tasks for knowledge enhanced pre-trained models with the base or large model size, such as ERNIE, ERNIE 2.0 and SpanBERT~\cite{joshi2020spanbert}, in which the base/large model size represent 12/24 layers Transformer respectively. In order to explore the effectiveness of knowledge enhanced large-scale pre-trained model, we propose the ERNIE 3.0 framework to pre-train model on massive unsupervised corpus including plain texts and knowledge graph. Furthermore, we employ various types of pre-training tasks to enable the model to learn the different levels of knowledge consisting of valuable lexical, syntactic and semantic information more effectively, in which the pre-training tasks spread three task paradigms, that is natural language understanding, natural language generation and knowledge extraction. Therefore, ERNIE 3.0 innovatively designs a \textbf{Continual Multi-Paradigms Unified Pre-training Framework} to enable the collaborative pre-training among multi-task paradigms. The explicit introduction of ERNIE 3.0 will be explained in the following sections.

\subsection{Overview of ERNIE 3.0 Framework}
The Framework of the ERNIE 3.0 is shown in Figure \ref{label:ernie 3.0 framework}, which can be widely used for pre-training, fine-tuning and zero/few-shot learning. Unlike the prevalent unified pre-training strategy of employing a shared Transformer network for different well-designed cloze tasks and utilizing specific self-attention masks to control what context the prediction conditions on, ERNIE 3.0 designs a new \textbf{\textit{Continual Multi-Paradigms Unified Pre-training Framework}}. We believed that the different task paradigms of natural language processing depend on identical underlying abstract features consistently, such as lexical information and syntactic information, but the requirements of top-level concrete features are incompatible, in which the natural language understanding tasks have the disposition to learn the semantic coherence while natural language generation tasks expect further contextual information. Therefore, inspired by the classical model architecture of multi-task learning, in which the lower layers are shared across all tasks while the top layers are task-specific, we proposed the ERNIE 3.0 to enable the different task paradigms to share the underlying abstract features learned in a shared network and utilizing the task-specific top-level concrete features learned in their own task-specific network respectively. Furthermore, in order to help the model efficiently learn the lexical, syntactic and semantic representations, ERNIE 3.0 exploits the continual multi-task learning framework introduced in ERNIE 2.0~\cite{sun2020ernie}. As for the application of different kinds of downstream tasks, we will first initialize the ERNIE 3.0 with the combination of parameters of a pre-trained shared network and corresponding task-specific networks for different task paradigms, and then execute the corresponding follow-up procedure using data from specific tasks.

We refer to the backbone shared network and task-specific networks as the \textbf{Universal Representation Module} and \textbf{Task-specific Representation Module}s in ERNIE 3.0. Specifically, the universal representation network plays the role of universal semantic features extractor (for example, it can be a multi-layer Transformer), in which the parameters are shared across all kinds of task paradigms, including natural language understanding, natural language generation and so on. And the task-specific representation networks undertake the function of extracting the task-specific semantic features, in which the parameters are learned by task-specific objectives. ERNIE 3.0 not only enables the model to distinguish the task-specific semantic information among different task paradigms, but also mitigates the dilemma that large-scale pre-trained models are difficult to implement with limited time and hardware resources, in which ERNIE 3.0 permits the models to only update the parameters of a task-specific representation network during the fine-tuning phase. Specifically, ERNIE 3.0 employs the collaborative architecture of a Universal Representation Module and two Task-specific Representation Modules, namely natural language understanding (NLU) specific representation module and natural language generation (NLG) specific representation module.

\subsubsection{Universal Representation Module}
ERNIE 3.0 uses a multi-layer Transformer-XL \cite{dai2019transformer} as the backbone network like other pre-trained models such as XLNet \cite{yang2019xlnet}, Segatron \cite{DBLP:journals/corr/abs-2004-14996} and ERNIE-Doc \cite{ding2020ernie}, in which Transformer-XL is similar to Transformer but introduces an auxiliary recurrence memory module to help modelling longer texts. We refer to the backbone as Universal Representation Module and it is shared across all the task paradigms. Proverbially, the Transformer can capture the contextual information for each token in the sequence via self-attention and generate a sequence of contextual embedding. It is evident that the larger the scale of Transformer model, the stronger its capacity to capture and store up various semantic information with different levels. Therefore, ERNIE 3.0 sets the universal representation module with a larger size to enable the model to effectively capture universal lexical and syntactic information from training data by learning various pre-training tasks of different paradigms. And what needs special attention is that the memory module is only valid for natural language generation tasks while controlling the attention mask matrices. 

\subsubsection{Task-specific Representation Module}
 Similar to the basic shared representation module, the task-specific representation module is also a multi-layer Transformer-XL, which is used to capture the top-level semantic representations for different task paradigms. ERNIE 3.0 sets the task-specific representation module to a manageable size, that is a base model size, instead of the multi-layer perceptron or shallow Transformer commonly used in multi-task learning, which will produce three obvious benefits, the first is that the base network has a stronger ability to capture semantic information than multi-layer perceptron and shallow Transformer; the second is that the task-specific networks with base model size enable ERNIE 3.0 to distinguish the top-level semantic information among different task paradigms without significantly increasing the parameters of a large-scale model; finally, the smaller model size of a task-specific network than a shared network would lead to realizable practical applications for large scale pre-trained model when only fine-tuning on the task-specific representation module. ERNIE 3.0 constructs two task-specific representation modules, that is NLU-specific representation module and NLG-specific representation module, in which the former is a  bi-directional modeling network while the latter is a uni-directional modeling network.

\subsection{Pre-training Tasks}
\label{sec:pre-training tasks}

We construct several tasks for various task paradigms to capture different aspects of information in the training corpora and make the capacity of understanding, generation and reasoning available to pre-trained model. 

\subsubsection{Word-aware Pre-training Tasks}\label{sec:Word-aware pretrain-task}

\textbf{Knowledge Masked Language Modeling} \,\, 
ERNIE 1.0~\cite{sun2019ernie} proposed an effective strategy to enhance representation through knowledge integration, namely Knowledge Integrated Masked Language Modeling task. It introduced phrase masking and named entity masking that predict the whole masked phrases and named entities to help the model learn the dependency information in both local contexts and global contexts. 

\textbf{Document Language Modeling} \,\, 
Generative pre-training models usually utilize traditional language model (such as GPT~\cite{radford2018improving}, GPT-2~\cite{radford2019language}) or sequence-to-sequence language model (such as BART~\cite{lewis2020bart}, T5~\cite{T5}, ERNIE-GEN~\cite{xiao2020erniegen}) as the pre-training task, the latter trains on the network with an auxiliary decoder structure. ERNIE 3.0 opt for traditional language model as the pre-training task to abate the network complexity and heighten the effectiveness of unified pre-training. In addition, to enable the NLG network of ERNIE 3.0 to model longer text, we introduce the Enhanced Recurrence Memory Mechanism proposed in ERNIE-Doc~\cite{ding2020ernie}, which can model a larger effective context length than traditional recurrence Transformer by changing the shifting-one-layer-downwards recurrence to the same-layer recurrence. 

\subsubsection{Structure-aware Pre-training Tasks}\label{sec:Structure-aware pretrain-task}
\textbf{Sentence Reordering} \,\, 
Sentence reordering task, which is introduced in ERNIE 2.0~\cite{sun2020colake}, aims to train the model to learn the relationship between sentences by reorganizing permuted segments. At length, a given paragraph is randomly split into 1 to m segments during pre-training and all of the combinations are shuffled by a random permuted order. Then, the pre-trained model is asked to reorganize these permuted segments, modeled as a k-class classification problem where \begin{math} k=\sum_{n=1}^{m} n!\end{math}.

\textbf{Sentence Distance} \,\, 
Sentence distance task, an extension of traditional next sentence prediction (NSP) task, is widely used in various pre-trained models to enhance their ability to learn the sentence-level information, which can be modeled as a 3-class classification problem. The three categories represent that the two sentences are adjacent, nonadjacent but in the same document and from two different documents respectively.

\begin{figure*}[ht]
	\centering
	\includegraphics[width=1.0\textwidth]{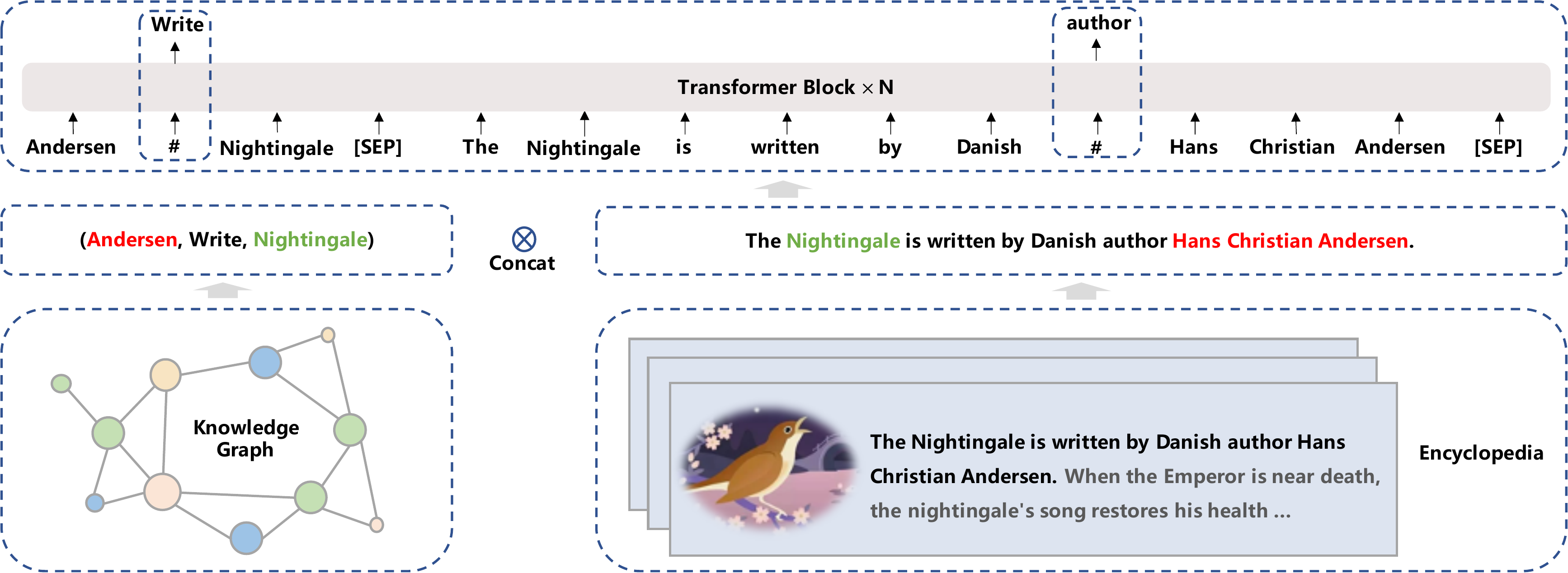}
	\caption{Universal Knowledge-Text Prediction.} 
\end{figure*}

\subsubsection{Knowledge-aware Pre-training Tasks}\label{sec: pretrain-task}
\textbf{Universal Knowledge-Text Prediction} \,\,
To incorporate knowledge into one pre-trained language model, we introduce universal knowledge-text prediction (UKTP) task, which is an extension of knowledge masked language modeling. While knowledge masked language modeling only requires unstructured texts, universal knowledge-text prediction task requires both unstructured texts and knowledge graphs. The universal knowledge-text prediction task is illustrated in Figure 2. Given a pair of triple from knowledge graph and the corresponding sentence from encyclopedia, we randomly mask relation in triple or words in a sentence. To predict the relation in the triple, the model needs to detect mentions of head entity and tail entity and determine semantic relationship that holds between them in the corresponding sentence. The essence of this process is similar to the distant supervision algorithm~\cite{mintz2009distant} in relation extraction tasks. The distant supervision algorithm assume that if two entities participate in a relation, any sentence that contain those two entities might express that relation. Meanwhile, to predict words in the corresponding sentence, the model not only considers the dependency information in the sentence, but also logical relationship in the triple. Specifically, the procedure of obtaining pairs of a triple and this corresponding sentence is as follows: given a document from encyclopedia, we first find the candidate triples in the knowledge graph whose mentions of head entity or tail entity is title of the document, and then select triples from candidate triples whose mentions of head entity and tail entity are mentioned in the same sentence in the document.

ERNIE 3.0 trains the NLU network through knowledge masked language modeling to improve the capacity of capturing the lexical information, trains the sentence reordering task and the sentence distance discerning task to strengthen the ability of capturing the syntactic information, and finally optimizes the model with the universal knowledge-text prediction task to improve knowledge memorization and reasoning. Meanwhile, ERNIE 3.0 trains the NLG network with the document language modeling task to enable various generation styles.

\subsection{Pre-training Process}
\subsubsection{Pre-training Algorithm}

Progressive training was originally proposed to improve stability, which starts from an efficient and small model and gradually increase the capacity \cite{simonyan2014very}. Recent study leverages this paradigm to accelerate model training. As large-scale pre-training keeps advancing the state-of-the-art(\cite{devlin2018bert}, \cite{radford2018improving}), their overwhelming computational consumption becomes the major burden towards further developing more powerful models(\cite{radford2019language}). Preliminary application of progressive training has been made on Transformer pre-training. BERT(\cite{devlin2018bert}) designs a two-stage training with a reduced sequence length for the first 90\% of updates. \cite{radford2019language} also gradually increase the batch size linearly from a small value to the full value. \cite{efficientnetv2} also notice that changing the regularization factors (e.g. \cite{srivastava2014dropout}, \cite{zhang2017mixup}) stage-wise with 
respect to the input size can speed up training networks. To further improve convergence speed of the training process, we propose to adjust the training regularization factors in a more comprehensive and smooth way by progressively and simultaneously increasing the training factors including the input sequence length, the batch size, the learning rate and the dropout rate. In fact, it is common that Transformer models adopts the learning rate warm-up strategy to increase training stability and our improved progressive learning strategy is compatible to the existing strategy.

\subsubsection{Pre-training Data}

\begin{table}[]
\centering
\resizebox{1\textwidth}{!}{
\begin{tabular}{@{}lccccccccccc@{}}
\toprule \toprule
\textbf{Corpus}       & \textbf{ERNIE 2.0} & \textbf{Search} & \textbf{Web}    & \textbf{QA-long} & \textbf{QA-short} & \textbf{Novel} & \textbf{Poetry\&Couplet} & \textbf{Medical} & \textbf{Law}   & \textbf{Fin}  & \textbf{KG}   \\ \midrule
\# of tokens & 17.8B     & 42.4B  & 314.7B & 33.8B   & 0.1B     & 96.4B & 46.5M           & 17.8B   & 16.2B & 0.6B & 0.7B \\
multiplier   & 20        & 7      & 1      & 3       & 40       & 1     & 20              & 1       & 1     & 10   & 10   \\ \midrule
\multicolumn{12}{l}{\# tokens of context length in each percentile using ERNIE-3.0 wordpiece tokenizer}                   \\ \midrule
50\%         & 135       & 75     & 793    & 184     & 15       & 2063  & 30              & 314     & 1162  & 843  & 16   \\
95\%         & 1257      & 827    & 2757   & 1168    & 22       & 3652  & 88              & 983     & 4587  & 1572 & 44   \\ \bottomrule \bottomrule
\end{tabular}}
\caption{Statistics of Pre-training Datasets.}
\label{tab:pre-dataset}
\end{table} 

 To ensure the success of the pre-training of ERNIE 3.0, we construct a large-scale, wide-variety and high-quality Chinese text corpora amounting to 4TB storage size in 11 different categories. To our best knowledge, this is currently the largest Chinese pre-training corpora compared with CLUECorpus2020~\cite{xu2020clue} (100GB), Chinese multi-modal pre-training data~\cite{lin2021m6} (300GB), WuDaoCorpus2.0 used by CPM-2~\cite{zhang2021cpm} (2.3TB Chinese data and 300GB English data) and PanGu Corpus~\cite{zeng2021pangu} (1.1TB). 
 
 In detail, we build the corpus for ERNIE 3.0 based on that from ERNIE 2.0 (including baike, wikipedia, feed and etc), Baidu Search (including Baijiahao, Zhidao, Tieba, Experience), Web text, QA-long, QA-short, Poetry~\footnote{\url{https://www.luge.ai/text-generation/chinese-poetry.html\#_1-chinese-poetry}}\&Couplet~\footnote{\url{https://github.com/v-zich/couplet-clean-dataset}}, Domain-specific data from medical, law and financial area and Baidu knowledge graph with more than 50 million facts. To improve the data quality, we adopt the following pre-processing strategies:
 \begin{itemize}
  \item Deduplication is conducted on different granularities including character level, paragraph level and document level. On the character level, we replace consecutive identical characters (i.e., spaces, tabs, exclamation mark, question mark and etc) with one single character. One the paragraph level, we replace two identical consecutive paragraphs consisting of $N$ sentences with one single paragraph where $0<N<100$. The two aforementioned deduplication strategies are critical for ERNIE 3.0 to generate non-repeating contents. At last, we adopted Message Digest Algorithm5 (MD5) to filter duplicate documents by comparing the sum of the MD5 of top-3 longest sentences from each document.
  \item Sentences with less than 10 words are filtered since they may be problematic or incomplete ones which contains limited semantic information for model pre-training.
  \item We further conduct sentence segmentation using regular expressions and word segmentation based on Baidu's word segmentation tool. This helps ERNIE 3.0 to learn better sentence boundary and named entity knowledge during pre-training.
 \end{itemize}
 Then, each dataset is multiplied by a user-defined multiplier number to increase the data diversity after truncating the data for NLU-network pre-training.

\subsubsection{Pre-training Settings}
Both the universal representation module and the task-specific representation modules of ERNIE 3.0 uses the Transformer-XL\cite{dai2019transformer} structure as the backbone. For the universal representation module, we adopt a structure with 48 layers, 4096 hidden units and 64 heads. For the task-specific representation modules, we adopt a structure with 12 layers, 768 hidden units and 12 heads. The total parameter of universal representation module and task-specific representation modules is 10 billion. The activation function used is GeLU\cite{hendrycks2016gaussian}.   
The maximum sequence length of context and the memory length of language generation is set to 512 and 128, respectively.  
The total batch size of all pre-training tasks is set to 6144.
We use Adam\cite{kingma2014adam} with learning rate of 1e-4, $\beta_1=0.9$, $\beta_2=0.999$, L2 weight decay of 0.01, learning rate warmup over the first 10,000 steps and linear decay of the learning rate.
In the first 10,000 steps, we also use the progressive learning to speedup convergence in the initial stage of pre-training. The model is trained for a total of 375 billion tokens with 384 NVDIA v100 GPU cards and is implemented on PaddlePaddle framework. 
By virtue of parameter sharding used in \cite{rajbhandari2020zero,ramesh2021zero}, we manage to reduce the memory usage of our model and address the problem of the total parameter of model exceeding the memory of a single GPU card.

\section{Experiments}
We compare the performance of ERNIE 3.0 with the state-of-the-art \footnote{the previous state-of-the-art results are all from the public single model that we can find.} pre-training models through fine-tuning on both natural language understanding tasks (in Sec.~\ref{sec:nlu-tasks}) and natural language generation tasks (in Sec.~\ref{sec:nlg-tasks}), and zero-shot learning (in Sec.~\ref{sec:zero-shot-tasks})\footnote{The previous SoTA results of ERNIE 2.0 and RoBERTa-wwm-ext on corresponding datasets are reproduced by ourselves, except for the datasets that already have released pre-trained results.}.

\subsection{Evaluation Tasks}\label{sec:evaluation-tasks}
We executed extensive experiments on 54 NLP tasks to evaluate the fine-tuning and zero-shot learning performances of the models.

\subsubsection{Natural Language Understanding Tasks}
45 datasets belonging to 14 kinds of natural language understanding tasks are used in our experiments, as follows:
\begin{itemize}
\item \textbf{Sentiment Analysis}: NLPCC2014-SC~\footnote{\url{http://tcci.ccf.org.cn/conference/2014/pages/page04_dg.html}}, SE-ABSA16\_PHNS~\footnote{\url{http://alt.qcri.org/semeval2016/task5/}}, SE-ABSA16\_CAME, BDCI2019~\footnote{\url{https://www.datafountain.cn/competitions/350}}.
\item \textbf{Opinion extraction}: COTE-BD~\cite{li2018cote}, COTE-DP~\cite{li2018cote}, COTE-MFW~\cite{li2018cote}.
\item \textbf{Natural Language Inference}: XNLI~\cite{conneau2018xnli}, OCNLI~\cite{xu2020clue}, CMNLI~\cite{xu2020clue}.  
\item \textbf{Winograd Schema Challenge} CLUEWSC2020~\cite{xu2020clue}.
\item \textbf{Relation Extraction}: FinRE~\cite{li2019finre}, SanWen~\cite{xu2017discourse}.
\item \textbf{Event Extraction}: CCKS2020~\footnote{\url{http://sigkg.cn/ccks2020/?page_id=69}}.
\item \textbf{Semantic Similarity}: AFQMC~\cite{xu2020clue}, LCQMC \cite{liu2018lcqmc}, CSL~\cite{xu2020clue}, PAWS-X~\cite{yang2019paws}, BQ Corpus~\cite{chen2018bq}. 
\item \textbf{Chinese News Classification}: TNEWS~\footnote{\url{https://github.com/aceimnorstuvwxz/toutiao-text-classfication-dataset}}, IFLYTEK~\cite{co2019iflytek}, THUCNEWS~\footnote{\url{http://thuctc.thunlp.org/}}, CNSE~\cite{liu2018matching}, CNSS~\cite{liu2018matching}.
\item \textbf{Closed-Book Question Answering}: NLPCC-DBQA~\footnote{\url{http://tcci.ccf.org.cn/conference/2016/dldoc/evagline2.pdf}}, CHIP2019, cMedQA~\cite{zhang2017cmedqa}, cMedQA2~\cite{zhang2018multi}, CKBQA~\footnote{\url{https://github.com/pkumod/CKBQA}}, WebQA~\cite{li2016webqa}.
\item \textbf{Named Entity Recognition}: CLUENER~\cite{xu2020clue}, Weibo~\cite{peng2015weibo}, OntoNotes~\cite{weischedel2011ontonotes}, CCKS2019~\footnote{\url{https://www.biendata.xyz/competition/ccks_2019_1/}}.
\item \textbf{Machine Reading Comprehension}: CMRC 2018 \cite{DBLP:journals/corr/abs-1810-07366}, CMRC2019~\cite{cui2020sentence}, DRCD \cite{shao2018drcd}, DuReader \cite{he2017dureader}, Dureader$_\text{robust}$~\cite{tang2020dureaderrobust}, Dureader$_\text{checklist}$, Dureader$_\text{yesno}$~\footnote{\url{https://aistudio.baidu.com/aistudio/competition/detail/49/?isFromLUGE=TRUE}}, C$^3$~\cite{sun2020c3}, CHID~\cite{zheng2019chid}.
% PD&CFT~\cite{cui2016pdcft},  CMRC2017~\cite{cui2017dataset}.
\item \textbf{Legal Documents Analysis}: CAIL2018-Task1~\cite{xiao2018cail2018}, CAIL2018-Task2~\cite{xiao2018cail2018}.
\item \textbf{Cant Understanding}: DogWhistle Insider, DogWhistle Outsider\cite{xu-etal-2021-blow}.
\item \textbf{Document Retrieval}: Sogou-log \cite{sogou-log}.
\end{itemize}

%\noindent\textbf{Natural Language Generation Tasks}:

\subsubsection{Natural Language Generation Tasks}
9 datasets belonging to 7 kinds of natural language generation tasks are used in our experiments, as follows:
\begin{itemize}
\item \textbf{Text Summarization}: LCSTS~\cite{hu2015lcsts}
\item \textbf{Question Generation}:KBQG~\footnote{\url{http://tcci.ccf.org.cn/conference/2017/dldoc/taskgline05.pdf}}, DuReader-QG~\cite{he2017dureader}, DuReader$_{\text{robust}}$-QG~\cite{tang2020dureaderrobust}.
\item \textbf{Closed-Book Question Answering}: MATINF-QA~\cite{xu2020matinf}.
\item \textbf{Math}: Math23K~\cite{wang2017math}.
\item \textbf{Advertisement Generation}: AdGen~\cite{shao2019adgen}.
\item \textbf{Translation}: WMT20-enzh~\cite{barrault2020wmt}.
\item \textbf{Dialogue Generation}: KdConv~\cite{zhou2020kdconv}.
\end{itemize}

\begin{table}[]
\small
\resizebox{1\textwidth}{!}{
\begin{tabular}{clcclcc}
\hline
\hline
\textbf{ID} & \textbf{Task}                                    & \textbf{Dataset}                             & \textbf{Metric}                  &                           & \textbf{Previous SoTA Model}       & \textbf{ERNIE 3.0}            \\ \hline
\multirow{5}{*}{1} & \multirow{5}{*}{Sentiment Analysis}          & NLPCC2014-SC                        & Acc.                    & \multicolumn{1}{l|}{Test} & 83.53 (SKEP)          & \textbf{86.00}       \\
                                       &      & \multicolumn{1}{l}{SE-ABSA16\_PHNS} & Acc.                    & \multicolumn{1}{l|}{Test} & 82.91 (SKEP)          & \textbf{93.95}       \\
                                    &         & SE-ABSA16\_CAME                     & Acc.                    & \multicolumn{1}{l|}{Test} & 90.06 (SKEP)          & \textbf{96.05}       \\
                                      &       & \multirow{2}{*}{BDCI2019}           & \multirow{2}{*}{Acc.}   & \multicolumn{1}{l|}{Dev}  & -                     & 96.83                \\
                                        &     &                                     &                         & \multicolumn{1}{l|}{Test} & 96.26 (ERNIE 2.0)     & \textbf{97.70}       \\ \hline
\multirow{3}{*}{2} & \multirow{3}{*}{Opinion Extraction}          & COTE-BD                             & F1                      & \multicolumn{1}{l|}{Test} & 84.50 (SKEP)          & \textbf{90.23}       \\
                                      &       & COTE-DP                             & F1                      & \multicolumn{1}{l|}{Test} & 86.30 (SKEP)          & \textbf{92.75}       \\
                                      &       & COTE-MFW                            & F1                      & \multicolumn{1}{l|}{Test} & 87.90 (SKEP)          & \textbf{89.90}       \\ \hline
\multirow{3}{*}{3} & \multirow{3}{*}{Natural Language Inference}  & OCNLI                               & Acc.                    & \multicolumn{1}{l|}{Dev}  & 78.80 (RoBERTa*)       & \textbf{82.75}       \\
                                       &      & \multirow{2}{*}{XNLI}               & \multirow{2}{*}{Acc.}   & \multicolumn{1}{l|}{Dev}  & 83.25 (Zen 2.0)       & \textbf{84.42}       \\
                                        &     &                                     &                         & \multicolumn{1}{l|}{Test} & 83.09 (Zen 2.0)       & \textbf{83.77}       \\ \hline
4 & Winograd Schema Challenge                    & WSC2020                             & Acc.                    & \multicolumn{1}{l|}{Dev}  & 69.70 (RoBERTa*)      & \textbf{95.40}       \\ \hline
\multirow{4}{*}{5} & \multirow{4}{*}{Relation Extraction}         & \multirow{2}{*}{FinRE}              & \multirow{2}{*}{F1}     & \multicolumn{1}{l|}{Dev}  & 63.33 (ERNIE 2.0)                   & \textbf{64.87}       \\
                                  &           &                                     &                         & \multicolumn{1}{l|}{Test} & 60.60 (ERNIE 2.0)     & \textbf{62.88}       \\
                                   &          & \multirow{2}{*}{SanWen}             & \multirow{2}{*}{F1}     & \multicolumn{1}{l|}{Dev}  & 79.92 (ERNIE 2.0)       & \textbf{81.32}       \\
                                    &         &                                     &                         & \multicolumn{1}{l|}{Test} & 77.97 (ERNIE 2.0)     & \textbf{82.59}       \\ \hline
\multirow{2}{*}{6} & \multirow{2}{*}{Event Extraction}            & \multirow{2}{*}{CCKS2020}           & \multirow{2}{*}{F1}     & \multicolumn{1}{l|}{Dev}  & 60.64 (ERNIE 2.0)                    & \textbf{61.70}       \\
                              &               &                                     &                         & \multicolumn{1}{l|}{Test} & 61.34 (ERNIE 2.0)     & \textbf{64.33}       \\ \hline
\multirow{8}{*}{7} & \multirow{8}{*}{Semantic Similarity}         & AFQMC                               & Acc.                    & \multicolumn{1}{l|}{Dev}  & 74.92 (RoBERTa*)      & \textbf{77.02}       \\
                   &                                             & \multirow{2}{*}{LCQMC}              & \multirow{2}{*}{Acc.}   & \multicolumn{1}{l|}{Dev}  & -                     & 90.29                \\
                   &                                             &                                     &                         & \multicolumn{1}{l|}{Test} & 89.16 (CPM-2)         & \textbf{90.38}       \\
                   &                                             & CSL                                 & Acc.                    & \multicolumn{1}{l|}{Dev}  & 82.17 (RoBERTa*)      & \textbf{84.50}       \\
                   &                                             & \multirow{2}{*}{PAWS-X}             & \multirow{2}{*}{Acc.}   & \multicolumn{1}{l|}{Dev}  & 86.25 (ERNIE 2.0)                     & \textbf{87.00}       \\
                   &                                             &                                     &                         & \multicolumn{1}{l|}{Test} & 86.35 (ERNIE 2.0)                  & \textbf{87.10}       \\
                   &                                             & \multirow{2}{*}{BQ Corpus}          & \multirow{2}{*}{Acc.}   & \multicolumn{1}{l|}{Dev}  & 87.11 (ZEN 2.0)                  & \textbf{87.41}       \\
                   &                                             &                                     &                         & \multicolumn{1}{l|}{Test} & 85.99 (ZEN 2.0)                  & \textbf{86.10}       \\ \hline
\multirow{8}{*}{8} & \multirow{8}{*}{Chinese News Classification} & TNEWS                               & Acc.                    & \multicolumn{1}{c|}{Dev}  & 58.32 (RoBERTa*)      & \textbf{69.94}       \\
                   &                          & IFLYTEK                             & Acc.                    & \multicolumn{1}{l|}{Dev}  & 62.75 (RoBERTa*)      & \textbf{63.45}       \\
                   &                          & \multirow{2}{*}{THUNCEWS}           & \multirow{2}{*}{Acc.}   & \multicolumn{1}{l|}{Dev}  & 97.7  (RoBERTa*)      & \textbf{98.33}       \\
                   &                          &                                     &                         & \multicolumn{1}{l|}{Test} & 97.6  (RoBERTa*)      & \textbf{98.66}       \\
                   &                          & \multirow{2}{*}{CNSE}               & \multirow{2}{*}{Acc.}   & \multicolumn{1}{l|}{Dev}  &   85.64  (RoBERTa*)                     & \textbf{88.94}                \\
                   &                          &                                     &                         & \multicolumn{1}{l|}{Test} & 85.57  (RoBERTa*)                  & \bf{88.92}           \\
                   &                          & \multirow{2}{*}{CNSS}               & \multirow{2}{*}{Acc.}   & \multicolumn{1}{l|}{Dev}  &     93.06 (ERNIE 2.0)               & \textbf{93.84}                \\
                   &                          &                                     &                         & \multicolumn{1}{l|}{Test} & 92.73 (ERNIE 2.0)                     & \bf{93.76}           \\ \hline
\multirow{7}{*}{9} & \multirow{7}{*}{Closed-Book Question Answering}          & \multirow{2}{*}{NLPCC-DBQA}         & \multirow{2}{*}{MRR/F1} & \multicolumn{1}{c|}{Dev}  & 96.04/85.69 (Zen 2.0) & \textbf{96.71/87.57} \\
                   &                          &                                     &                         & \multicolumn{1}{l|}{Test} & 96.11/86.47 (Zen 2.0) & \textbf{96.50/88.49} \\
                   &                          & CHIP2019                            & Acc.                    & \multicolumn{1}{l|}{Test} & 89.22 (ERNIE 2.0)     & \textbf{89.90}       \\ 
                   &                          & \multirow{2}{*}{cMedQA}             & \multirow{2}{*}{Acc.}   & \multicolumn{1}{l|}{Dev}  & 78.6 (BERT\_BiGRU*)    & \textbf{84.60}       \\
                   &                          &                                     &                         & \multicolumn{1}{l|}{Test} & 78.2 (BERT\_BiGRU*)    & \textbf{82.65}       \\
                   &                          & \multirow{2}{*}{cMedQA2}            & \multirow{2}{*}{Acc.}   & \multicolumn{1}{l|}{Dev}  & 81.3 (BERT\_BiGRU*)    & \textbf{83.48}       \\
                   &                          &                                     &                         & \multicolumn{1}{l|}{Test} & 82.2 (BERT\_BiGRU*)    & \textbf{83.68}       \\ \hline
\multirow{6}{*}{10}& \multirow{6}{*}{Named Entity Recognition}    & CLUENER                             & F1                      & \multicolumn{1}{c|}{Dev}  & 80.42 (RoBERTa*)      & \textbf{81.23}       \\
                   &                          & \multirow{2}{*}{Weibo}              & \multirow{2}{*}{F1}     & \multicolumn{1}{l|}{Dev}  & -            & 70.06                    \\
                   &                         &                                     &                         & \multicolumn{1}{l|}{Test} & 67.60 (Glyce+BERT)    & \textbf{69.23}   \\
                   &                          & \multirow{2}{*}{OntoNotes}          & \multirow{2}{*}{F1}     & \multicolumn{1}{l|}{Dev}  & -                     & 79.59            \\
                   &                          &                                     &                         & \multicolumn{1}{l|}{Test} & 81.63 (Glyce+BERT)    & \textbf{82.64}                    \\
                   &                          & CCKS2019                            & F1                      & \multicolumn{1}{l|}{Test} & 81.58 (ERNIE 2.0)     & \textbf{82.70}       \\ \hline
\multirow{4}{*}{11}& \multirow{4}{*}{Cant Understanding}          & \multicolumn{1}{l}{\multirow{2}{*}{DogWhistle Insider}}  & \multirow{2}{*}{Acc.}  & \multicolumn{1}{c|}{Dev}  & 75.4 (ALBERT)      & \textbf{79.06}       \\
                   &                          & \multicolumn{1}{l}{}                                     &       & \multicolumn{1}{l|}{Test} & 76.1 (ALBERT)              & \textbf{79.22}       \\
                   &                          & \multicolumn{1}{l}{\multirow{2}{*}{DogWhistle Outsider}} & \multirow{2}{*}{Acc.}      & \multicolumn{1}{l|}{Dev}  & 34.6 (ALBERT)        & \textbf{38.68}       \\
                   &                          & \multicolumn{1}{l}{}                                     &                            & \multicolumn{1}{l|}{Test} & 34.6 (ALBERT)        & \textbf{38.22}       \\ \hline\hline
\end{tabular}
}
\end{table}

\begin{table}[]
\resizebox{1\textwidth}{!}{
\begin{tabular}{clcclcc}
\hline\hline
\textbf{ID} & \textbf{Task}                                         & \textbf{Dataset}                             & \textbf{Metric}    &                                 & \textbf{Previous SoTA Model}         & \textbf{ERNIE 3.0}            \\ \hline
\multirow{14}{*}{12} & \multirow{14}{*}{Machine  Reading  Comprehension} & CMRC2018                            & EM/F1                              & \multicolumn{1}{c|}{Dev}  & 74.3/90.5 (ERNIE-Gram)  & \textbf{75.30/92.29} \\
                     &                             & CRMC2019                            & QAC/PAC                            & \multicolumn{1}{c|}{Dev}  & 82.6/23.3 (RoBERTa*)     & \textbf{92.53/57.33} \\
                     &                             & \multirow{2}{*}{DRCD}               & \multirow{2}{*}{EM/F1}             & \multicolumn{1}{c|}{Dev}  & 90.8/95.3 (MacBERT)     & \textbf{91.54/96.45} \\
                     &                             &                                     &                                    & \multicolumn{1}{c|}{Test} & 90.9/95.3 (MacBERT)     & \textbf{91.41/95.84} \\
                     &                             & DuReader                            & EM/F1                              & \multicolumn{1}{c|}{Dev}  & 64.2/77.3 (ERNIE 2.0)   & \textbf{67.69/79.66} \\
                     &                             & \multirow{2}{*}{DuReader$_\text{robust}$}    & \multirow{2}{*}{EM/F1}             & \multicolumn{1}{c|}{Dev}  & 75.23/86.77 (ERNIE 2.0)  & \textbf{77.27/88.54}\\
                     &                             &                                     &                                    & \multicolumn{1}{c|}{Test} & 51.20/67.96 (ERNIE 2.0)  & \textbf{60.87/75.63} \\
                     &                             & \multirow{2}{*}{DuReader$_\text{checklist}$} & \multirow{2}{*}{EM/F1}             & \multicolumn{1}{c|}{Dev}  & 55.66/64.12 (ERNIE 2.0) & \textbf{61.33/70.59} \\
                     &                             &                                     &                                    & \multicolumn{1}{c|}{Test} & 59.11/48.79 (ERNIE 2.0) & \textbf{64.87/53.82} \\
                     &                             & \multirow{2}{*}{DuReader$_\text{yesno}$}   & \multirow{2}{*}{Acc.}              & \multicolumn{1}{c|}{Dev}  & 88.69 (ERNIE 2.0)       & \textbf{89.95}       \\
                     &                             &                                     &                                    & \multicolumn{1}{c|}{Test} & 88.82 (ERNIE 2.0)       & \textbf{89.64}       \\  
                     &                             & \multirow{2}{*}{C3}                 & \multirow{2}{*}{Acc.}              & \multicolumn{1}{c|}{Dev}  & -                       & 87.63                \\
                     &                             &                                     &                                    & \multicolumn{1}{c|}{Test} & 86.1 (CPM-2)            & \textbf{86.69}       \\
                     &                             & CHID                                & Acc.                               & \multicolumn{1}{c|}{Dev}  & 85.81 (RoBERTa*)        & \textbf{91.67}       \\ \hline
\multirow{4}{*}{13}  & \multirow{4}{*}{Legal Document Analysis}                 & \multirow{2}{*}{CAIL2018 Task1}     & \multirow{2}{*}{F1-macro/F1-micro} & \multicolumn{1}{c|}{Dev}  & 83.85/91.50 (ERNIE 2.0) & \textbf{88.64/93.11} \\
                     &                             &                                     &                                    & \multicolumn{1}{c|}{Test} & 80.40/89.94 (ERNIE 2.0) & \textbf{86.83/91.82} \\
                     &                             & \multirow{2}{*}{CAIL2018 Task2}     & \multirow{2}{*}{F1-macro/F1-micro} & \multicolumn{1}{c|}{Dev}  & 78.58/89.46 (ERNIE 2.0) & \textbf{82.62/90.93} \\
                     &                             &                                     &                                    & \multicolumn{1}{c|}{Test} & 75.35/86.97 (ERNIE 2.0) & \textbf{81.10/88.52} \\

\hline
\multirow{1}{*}{14}  & \multirow{1}{*}{Document Retrieval}                 & \multirow{1}{*}{Sogou-log}     & \multirow{1}{*}{MRR/NDCG@1} & \multicolumn{1}{c|}{Test}  & 36.3/35.5 (CPM-2) & \textbf{38.20/37.24} \\
                      \hline\hline

\end{tabular}
}
\caption{Results on Natural Language Understanding Tasks. We compare ERNIE 3.0 with 10 previous SoTA baselines including CPM-2\cite{zhang2021cpm}, ERNIE 2.0\cite{sun2020ernie}, ERNIE-Gram\cite{xiao2020erniegram}, SKEP\cite{tian2020skep}, RoBERTa-wwm-ext-large\cite{cui2019pre} (marked as RoBERTa*), ALBERT\cite{lan2019albert}, MacBERT\cite{cui2020revisiting}, Zen 2.0\cite{song2021zen}, Glyce\cite{meng2019glyce} and crossed BERT siamese BiGRU\cite{cui2020chinese} (marked as BERT\_BiGRU*).}\label{tab:result-understanding}
\end{table}

\subsection{Experiments on Fine-tuning Tasks}\label{sec:finetune-tasks}
\subsubsection{Fine-tuning on Natural Language Understanding Tasks}\label{sec:nlu-tasks}

The results of natural language understanding tasks are reported in Table~\ref{tab:result-understanding}.

\noindent\textbf{Sentiment Analysis}. Sentiment Analysis is a classification task aiming to determine whether a sentence is positive, negative, or neutral. We consider 4 datasets from different domains, including shopping (NLPCC2014-SC), electronics (SE-ABSA16\_PHNS, SE-ABSA16\_CAM), and financial (BDCI2019). ERNIE 3.0 achieves a substantial improvement on all four datasets.

\noindent\textbf{Opinion Extraction}. Similar to the sentiment analysis task, opinion extraction requires the model to mine the opinion of a sentence. We use 3 sub-datasets from Chinese Customer Review (COTE). Experiment results show that ERNIE 3.0 also outperforms the current SoTA system by a great margin.

\noindent\textbf{Natural Language Inference}. Natural Language Inference is the task to determine whether a given premise semantically entails another hypothesis. We use OCNLI and XNLI datasets. The results indicate that ERNIE 3.0 has achieved 3.9 and 0.7 accuracy improvement on two datasets, respectively. The improvement on the XNLI dataset is quite limited, and this may be due to the poor quality of the dataset since the XNLI dataset is translated from English.

\noindent\textbf{Winograd Schemas Challenge}. WSC2020 is an anaphora resolution task where the model is asked to decide whether a pronoun and a noun in a sentence co-refer, ERNIE 3.0 achieves a significant improvement of 25.7 points.

\noindent\textbf{Relation Extraction}. The task of relation extraction is to identify the relationship between different entities like persons and organizations. We consider FinRE and SanWen -- two relation extraction datasets for financial news and Chinese literature respectively. ERNIE 3.0 outperforms the previous SoTA model by 2.46 points on average.

\noindent\textbf{Event Extraction}. Similar to relation extraction, the event extraction task aims to identify the event entities and classify them into different categories. We choose CCKS2020 -- a text-level event subject extraction dataset of financial field. ERNIE 3.0 has 3 points of improvement on the test set.

\noindent\textbf{Semantic Similarity}. Semantic Similarity is a classic NLP task that determines the similarity between various terms such as words, sentences, documents. In this work, we focus on sentence-level similarity tasks. We test ERNIE 3.0 on several datasets in varied fields including AFQMC, LCQMC, CSL, PAWS-X, and BQ.  Experiment results show that ERNIE 3.0 outperforms the baseline models by a remarkable margin. Especially, under comparable number of parameters, ERNIE 3.0 surpasses CPM-2 with 1.2 points on LCQMC dataset.

\noindent\textbf{Chinese News Classification}. We also evaluate ERNIE 3.0 on Chinese news classification. We consider 6 datasets including news title (TNEWS), app descriptions (IFLYTEK), and news stories (THUCNEWS, CNSE, CNSS). Under different types of classification tasks, ERNIE 3.0 can consistently achieve better accuracy with 2.8 points improvement on average.

\noindent\textbf{Closed-Book Question Answering}. Closed-Book Question Answering aims to directly answer the questions without any additional references or knowledge. We select a general QA dataset NLPCC-DBQA and three medical field datasets -- CHIP2019,  cMedQA, and cMedQA2 to test the ability of ERNIE 3.0. Experiment results show that  ERNIE 3.0 performs better on all QA tasks, we believe knowledge enhanced pre-training methods do bring benefits to the closed-book QA task.

\noindent\textbf{Cant Understanding}. Cant, also known as doublespeak, is an advanced language usage for humans. However, it is rather difficult for machines to understand this type of language. We test the cant understanding ability of ERNIE 3.0 on DogWhistle -- a dataset based on \emph{Decrypto} game.  The model is required to select the right answer with the guidance of the corresponding cant. ERNIE 3.0 gets the best result and shows its potential for understanding much more difficult languages.

\noindent\textbf{Named Entity Recognition}. Named Entity Recognition is a classical NLP task of extracting and classifying entities in text. We select widely used OntoNotes, CLUENER, Weibo, and a domain-specific dataset CCKS2019. From the results, ERNIE 3.0 performs better than the baseline models across all datasets.

\noindent\textbf{Machine Reading Comprehension}. We comprehensively evaluate the ability of ERNIE 3.0 on machine reading comprehension in different aspects, including span-predict reading comprehension (CMRC2018, DuReader, DRCD, DuReader$_\text{checklist}$),  multiple-choice reading comprehension (C3, DuReader$_\text{yesno}$), cloze and completion (CHID, CMRC2019), and robustness test (Dureader$_\text{robust}$). With the help of knowledge enhanced pre-training, ERNIE 3.0 surpasses the baseline models with significant enhancements on all types of tasks. To be more specific, ERNIE 3.0 achieve at least 1.0 points of EM improvement on 5 span-predict tasks and 0.89 accuracy improvement on multiple-choice tasks on average. Also, under comparable number of parameters, ERNIE 3.0 outperforms CPM-2 with 0.6 points on C3 dataset. For the robustness test, ERNIE 3.0 also performs best on the test set with over-sensitivity and over-stability samples.

\noindent\textbf{Legal Documents Analysis}. Next, we test the ability of ERNIE 3.0 on document analysis, we choose two domain-specific tasks of law. These two datasets from CAIL2018 are both multi-label document classification tasks. ERNIE 3.0 outperforms ERNIE 2.0 with remarkable increment.

\noindent\textbf{Document Retrieval}. Document retrieval aims to match documents given queries. We evaluate the retrieval ability of ERNIE 3.0 on Sogou-Log. Following previous work \cite{zhang2021cpm}, we report NDCG@1 performance on the test-same test set and MRR performance on the test-raw test set and ERNIE 3.0 outperforms CPM-2.

\subsubsection{Fine-tuning on Natural Language Generation Tasks}\label{sec:nlg-tasks}
The results of natural language generation tasks are reported in Table~\ref{tab:result-generation}.

\noindent\textbf{Text Summarization}. 
We consider a Large Scale Chinese Short Text Summarization (LCSTS) dataset which requires a model to understand the text and refine the key information to generate
coherent, informative summaries. LCSTS is a classic
Chinese text summarization dataset which consists of 2 million real Chinese short texts with short summaries from Sina Weibo. 
ERNIE 3.0 achieves 48.46\% Rouge-L score which outperforms CPM-2 with comparable number of parameters (11B) and current SoTA ProphetNet-zh.

\noindent\textbf{Question Generation}. Question Generation is the reverse task of Machine Reading Comprehension (MRC) which requires the model to understand a document and generate a
reasonable question based on a given short answer. We use a suite of three datasets including knowledge base question generation (KBQG), two MRC datasets named Dureader and Dureader${_\text{robust}}$. ERNIE 3.0 performs best on these three datasets compared to the baselines.

\noindent\textbf{Math}. To test ERNIE 3.0's ability to perform simple arithmetic operations, we consider the Math23K dataset which contains 23,161 real math word problems for elementary school students with problem descriptions, structured equations and answers. ERNIE 3.0 is fine-tuned to generate the postfix expression of the structured equation given the problem description, then the final answer can be calculated using the Python \textit{eval()} function (note that the `[' and `]' should be replaced with `(' and `)' respectively, also the `\%' should be replaced with `*0.01' to avoid the failed solutions using Python \textit{eval()} function). It shows that ERNIE 3.0 is a great math solver which achieves high accuracy 75\% compared to CPM-2 69.37\%.

\begin{table}[]
\centering
\resizebox{\textwidth}{!}{%
\begin{tabular}{@{}llccccccc@{}}
\toprule \toprule
\textbf{Task}                            & \textbf{Dataset}            & \textbf{Metric}                       & \textbf{RoBERTa-Large} & \textbf{ERNIE 2.0-Large} & \textbf{ProphetNet-zh} & \textbf{mT5}   & \textbf{CPM-2} & \textbf{ERNIE 3.0}      \\ \midrule
Text Summarization  & LCSTS              & \multicolumn{1}{c|}{ROUGE-L} & 40.98   & 41.38     & 37.08         & 34.8  & 35.88 & \textbf{48.46} \\ \midrule
\multirow{3}{*}{Question Generation} & KBQG               & \multicolumn{1}{c|}{BLEU-4}  & -       & 57.40     & -             & -     & -     & \textbf{64.70} \\
                                     & DuReader-QG        & \multicolumn{1}{c|}{BLEU-4}  & 32.29   & 34.15     & -             & -     & -     & \textbf{48.36} \\
                                     & DuReader$_\text{robust}$-QG & \multicolumn{1}{c|}{BLEU-4}  & 37.10   & 39.30     & -             & -     & -     & \textbf{41.70} \\ \midrule
Closed-Book Question Answering       & MATINF-QA          & \multicolumn{1}{c|}{ROUGE-L} & -       & -         & 15.47         & -     & -     & \textbf{17.33} \\ \midrule
Math                                 & Math23K            & \multicolumn{1}{c|}{Acc.}     & -       & -         & -             & 61.60 & 69.37 & \textbf{75.00}          \\ \midrule
Advertisement Generation             & AdGen              & \multicolumn{1}{c|}{BLEU-4}  & -       & -         & -             & 9.82  & 10.60 & \textbf{30.16} \\ \midrule
Translation                          & WMT20-enzh         & \multicolumn{1}{c|}{BLEU}    & -       & -         & -             & 23.98 & 26.21 &     \textbf{26.80}           \\ \midrule
Dialogue Generation                  & KdConv             & \multicolumn{1}{c|}{BLEU-4}                    & 15.75   &   13.94        & -             & -     & -     & \textbf{23.85}          \\ \bottomrule \bottomrule
\end{tabular}%
}
\caption{Results on Natural Language Generation Tasks. We reported the results on the test set.}
\label{tab:result-generation}
\end{table}

\noindent\textbf{Advertisement Generation}. We consider AdGen which consists of 119K pairs of advertising text and clothing specification tables from a Chinese e-commerce platform.
It requires the model to generate a long advertising text that covers all given attribute-value pairs for a piece of clothing.
An attribute-value pair is joined with a colon, and several attribute-value pairs are concatenated sequentially using a `|' according to their segment number.
Then we take the structural attribute-value pairs string as input for ERNIE 3.0. It shows that ERNIE 3.0 is capable to generate a coherent and intriguing long advertising text by extracting information from a structural input with 19.56 percent point improvement w.r.t BLEU-4 compared to CPM-2.

\noindent\textbf{Translation}. For ERNIE 3.0, we mainly consider the pre-training on Chinese corpus. To test its multilingual ability, we expand our vocabulary to include extra 10K English subwords. On a classic multilingual dataset WMT20-enzh, we fine-tuned ERNIE 3.0 to translate English to Chinese. 
Compared to mT5-xxLarge and CPM-2, ERNIE 3.0~\footnote{Due to the large size of the training dataset of WMT20-enzh, ERNIE 3.0 is not fully trained to convergence. We reported the BLEU score at 1.5 epoch checkpoint using SacreBLEU project~\cite{post-2018-call}.} is the best and presents superior multilingual ability.

\noindent\textbf{Dialogue Generation}. Next, we evaluate ERNIE 3.0 on Dialog Generation task. We consider a Chinese multi-domain knowledge-driven conversation dataset that contains
4.5K conversations from three domains (film, music, and travel). We train and test ERNIE 3.0 on the fused set of data from aforementioned three domains by only giving dialogue history to generate the current utterance. 
Knowledge triplets are excluded from inputs, so it's suitable to test a model's ability to model multi-turn conversations by leveraging inherent knowledge during pre-training.
Compared to baselines, ERNIE 3.0 improves the performance a lot by 8.1 percent point, and we believe the knowledge graph enhanced pre-training attributes a lot.

% \usepackage{booktabs}
% \usepackage{multirow}
% \usepackage{graphicx}

\begin{table}[]
\centering
\resizebox{\textwidth}{!}{%
\begin{tabular}{@{}lllcccc@{}}
\toprule \toprule
\textbf{Task Paradigm}        & \textbf{Task}                 & \textbf{Dataset}   & \textbf{Metric} & \textbf{RoBERTa-Large} & \textbf{ERNIE 2.0-Large} & \textbf{ERNIE 3.0}   \\ \midrule
\multirow{3}{*}{\textbf{NLU}} & Sentiment Analysis            & NLPCC14-SC         & Acc.        & 83.56                  & 84.36                    & \textbf{86.00}       \\
                              & Machine Reading Comprehension & DuReader$_\text{robust}$    & EM/F1           & 51.10/67.18            & 51.20/67.96              & \textbf{60.87/75.63} \\
                              & Semantic Similarity           & LCQMC              & Acc.        & 87.40                  & 87.90                    & \textbf{90.38}       \\ \midrule
\multirow{3}{*}{\textbf{NLG}} & Question Generation           & DuReader$_\text{robust}$-QG & BLEU-4          & 37.10                  & 39.30                    & \textbf{41.70}       \\
                              & Text Summarization            & LCSTS              & Rouge-L         & 40.98                  & 41.38                    & \textbf{48.46}       \\
                              & Dialogue Generation           & KdConv             & BLEU-4          & 15.75                  & 13.94                    & \textbf{23.85}       \\ \midrule
\multicolumn{4}{l}{\textbf{Average}}                                                                 & 53.99                  & 54.41                    & \textbf{59.77}       \\ \bottomrule \bottomrule
\end{tabular}%
}
\caption{Results on the LUGE benchmark. We reported the results on the test set.}
\label{tab:result-luge}
\end{table}

\begin{table}[]
\centering
\resizebox{\textwidth}{!}{%
\begin{tabular}{@{}lcccccc@{}}
\toprule \toprule
\textbf{Task Type}                              & \textbf{Dataset} & \textbf{Metric}            & \textbf{CPM-1} & \textbf{PanGu-$\alpha$-2.6B} & \textbf{PanGu-$\alpha$-13B} & \textbf{ERNIE 3.0}   \\ \midrule
\multirow{2}{*}{Chinese News Classification}    & TNEWS            & \multicolumn{1}{c|}{Acc.}   & 65.44          & 60.95               & 60.26              & \textbf{68.40}       \\
                                                & IFLYTEK          & \multicolumn{1}{c|}{Acc.}   & 68.91          & 74.26               & 73.80              & \textbf{75.34}       \\ \midrule
\multirow{2}{*}{Semantic Similarity}            & AFQMC            & \multicolumn{1}{c|}{Acc.}   & 66.34          & 59.29               & 65.76              & \textbf{68.99}       \\
                                                & CSL              & \multicolumn{1}{c|}{Acc.}   & 52.30          & 50.50               & 49.30              & \textbf{55.63}       \\ \midrule
\multirow{2}{*}{Natural Language Inference}     & OCNLI            & \multicolumn{1}{c|}{Acc.}   & 44.20 & 42.61               & 41.53              & \textbf{44.31}                \\
                                                & CMNLI            & \multicolumn{1}{c|}{Acc.}   & 49.10          & 47.56               & 49.29     & \textbf{49.41}                \\ \midrule
Winograd Schema Challenge                       & WSC2020          & \multicolumn{1}{c|}{Acc.}   & 73.68          & 73.36               & 75.00              & \textbf{78.38}       \\ \midrule
\multirow{6}{*}{Cloze and completion}           & CHID             & \multicolumn{1}{c|}{Acc.}   & 68.62          & 68.73               & 70.64              & \textbf{77.78}       \\
                                                & PD               & \multicolumn{1}{c|}{Acc.}   & 35.73          & 38.47               & 43.84              & \textbf{66.07}       \\
                                                & CFT              & \multicolumn{1}{c|}{Acc.}   & 38.99          & 42.39               & 46.60              & \textbf{49.30}                    \\
                                                & CMRC2017         & \multicolumn{1}{c|}{Acc.}   & 24.60          & 37.83               & 38.90              & \textbf{56.66}       \\
                                                & CMRC2019         & \multicolumn{1}{c|}{Acc.}   & 47.69          & 61.93               & 68.19              & \textbf{75.00}                \\
                                                & WPLC             & \multicolumn{1}{c|}{PPL}   & -              & 48.98               & 45.85              & \textbf{17.03}       \\ \midrule
\multirow{4}{*}{Machine Reading Comprehension}  & C3               & \multicolumn{1}{c|}{Acc.}   & 49.81          & 53.42               & \textbf{54.47}              & 52.62                    \\
                                                & CMRC2018         & \multicolumn{1}{c|}{EM/F1} & 0.59/10.12     & 1.21/16.65          & 1.46/19.28         & \textbf{7.61/25.61}  \\
                                                & DRCD             & \multicolumn{1}{c|}{EM/F1} & 0.00/4.62      & 0.80/9.99           & 0.66/10.55         & \textbf{10.58/26.29} \\
                                                & DuReader         & \multicolumn{1}{c|}{EM/F1} & 16.63          & 21.07               & 24.46              & \textbf{29.79}       \\ \midrule
\multirow{2}{*}{Closed-book Question Answering} & WebQA            & \multicolumn{1}{c|}{EM/F1} & 6.00/12.59     & 4.43/13.71          & 5.13/14.47         & \textbf{22.53/38.95}          \\
                                                & CKBQA            & \multicolumn{1}{c|}{Acc.}   & 13.40          & 14.61               & 14.21              & \textbf{20.64}       \\ \bottomrule \bottomrule
\end{tabular}%
}
\caption{Results on zero-shot learning tasks.}
\label{tab:zero-shot-result}
\end{table}

\subsubsection{LUGE benchmark}\label{sec:luge-tasks}

In order to further evaluate the capabilities of different models comprehensively and conveniently, we conduct experiments on the Language Understanding and Generation Evaluation Benchmarks(LUGE))~\footnote{\url{https://www.luge.ai/}}. We use six representative tasks (see Tab.~\ref{tab:result-luge}) from LUGE. ERNIE 3.0 delivers an average 5.36 percent improvement over leading pre-trained models such as ERNIE 2.0 and RoBERTa.

%s LUGE provides a easy-to-use and reproducible environment for all users. Datasets and evaluation scripts can be downloaded to train and evaluate the models offline. Users could submit the result on test dataset and make fair comparison with current models.

\subsection{Experiments on Zero-shot Learning}\label{sec:zero-shot-tasks}
We have demonstrated that ERNIE 3.0 is superior to previous SoTA methods on both NLU and NLG tasks following the pretraining-then-finetuning 
paradigm. In this section, we conduct various types of tasks with the zero-shot setting where a model is applied without any gradient updates
or fine-tuning. ERNIE 3.0 achieves strong performance compared to recently proposed large-scale language models such as CPM-1 (2.6B), PanGu-$\alpha$-2.6B
and PanGu-$\alpha$-13B on most downstream tasks. 
At last, we show that ERNIE 3.0 can generate more coherent, natural and accurate responses rated on our manually collected 450 cases across 13 different tasks.

\subsubsection{Evaluation}
The evaluation methods can be classified into two categories, namely perplexity-based method and generation-based method. 

\begin{itemize}
    \item \textbf{Perplexity-based Method}. On tasks that choose one single correct answer from multiple candidates such as CHID and CMRC2017, we compare the per-token perplexity score~\footnote{The perplexity score of a sample is normalized by the number of tokens.} when filling each answer into the blank of the context. The one with lower per-token perplexity score will be the predicted as the correct answer. On tasks that require binary or multiple classification, we assign each label with a more semantically meaningful name and use a prompt to formalize the context and the label as a human-readable text. Then, this kind of tasks can be treated as multi-choice tasks. The prompts we used are similar to that in CPM-1 and PanGu-$\alpha$.

\item \textbf{Generation-based Method}. On tasks with free-form completion such as Closed-book QA, we use beam search with a beam width of 8 and no length penalty. The maximum generated length of a completion is limited by a pre-defined number based on 95\% percentile point of answers' length on the dataset. Then metrics such as exact match (EM), F1 and Rouge-1 are used. On tasks with restrained completion such as extractive MRC, we use restrained beam search with the same parameters as before. A Trie-Tree is constructed for each sample to
efficiently and effectively restrain the space of generation and only generate completion occurred in a given text.
\end{itemize}

\subsubsection{Results}

% \usepackage{booktabs}
% \usepackage{multirow}
% \usepackage{graphicx}

\noindent\textbf{Chinese News Classification}. For the TNEWS and IFLYTEK datasets, there are 15 and 119 categories respectively. We randomly sample three candidates as negative labels for each sample and compare the per-token perlexity score among these four choices. 
This sampling strategy is aligned with CPM-1's and PanGu-$\alpha$'s to reduce the total computational cost since we need to calculate per-token perlexity score 
for each candidate separately. ERNIE 3.0 performs well on TNEWS even reaching competitiveness with prior state-of-the-art fine-tuning approaches and performs slightly well on IFLYTEK.

\noindent\textbf{Semantic Similarity}. We consider AFQMC and CSL datasets. ERNIE 3.0 outperforms baselines at a large margin. 
However, the accuracy is slightly above than a random-guess model. 
This may be partly attributed to the sub-optimal selection of the prompt (like \textsc{The following two sentences have the same/different semantics: \$SENT\_A. \$SENT\_B.}).

\noindent\textbf{Natural Language Inference}. ERNIE 3.0 is evaluated on two NLI datasets, namely OCNLI and CMNLI where CMNLI consists of XNLI and MNLI by translating English to Chinese. We use the prompt as \textsc{\$SENT\_A? No/Yes/Maybe, \$SENT\_B}. The performance of ERNIE 3.0 is comparable to baselines, it shows that there is still a large room for improvement for pre-trained models on zero-shot NLI task.

\noindent\textbf{Winograd Schema Challenge}: We formalize the WSC2020 dataset as a multi-choice completion task where a pronoun is replaced with each candidates to calculate the per-token perplexity of a sample. ERNIE 3.0 improves the performance by 3.38 percent point compared to PanGu-$\alpha$-13B. 

\noindent\textbf{Cloze and completion}. On the CHID dataset, we split each sentence that contains only one blank word as a sample, and formalize as a multi-choice task. ERNIE 3.0 achieves the best score among baselines. For Chinese Word Prediction with Long Context (Chinese WPLC), a sample consists of a masked text and a correct word. Following PanGu-$\alpha$, we replace the mask token with the correct word and calculate the perplexity score of a whole sentence. Compared to PanGu-$\alpha$, ERNIE 3.0 achieves much lower perplexity score. On the CMRC2019 dataset, we randomly sample three negative candidates for each blank from the original candidates, then beam search is applied to calculate the optimal path for a sample. We also formalize the PD, CFT and CMRC2017 as multi-choice tasks where the text before the blank is taken as the input, and the multiple choices are the words the appear in the whole text. ERNIE 3.0 surpassed the baselines with a large margin. 

\noindent\textbf{Machine Reading Comprehension}. We consider four MRC datasets. On C3, a multi-choice machine reading comprehension tasks, 
we use the prompt as \textsc{Question: \$Question? Answer: \$Choice. The answer is in the following document: \$Document}. For CMRC2018, DRCD and DuReader, we evaluate ERNIE 3.0 using generation-base method and the prompt is \textsc{Document: \$Document. Question: \$Question? Answer:}. 
ERNIE 3.0 outperforms baselines with a large margin on CMRC2018, DRCD and DuReader dataset.

\noindent\textbf{Closed-book Question Answering}. We evaluated ERNIE 3.0 on two Closed-book Question Answering datasets which require the model to generate answers using its inherent knowledge learned during pre-training. WebQA is a large scale real-word QA dataset from Baidu Zhidao. We only provide ERNIE 3.0 with the question without additional evidence. The prompt is similar to MRC's but without document input (\textsc{Question: \$Question? Answer:}). ERNIE 3.0 achieves better performance compared to baselines. We presented the detailed analysis about CKBQA dataset in Section.~\ref{sec:analysis}.

\subsubsection{Case Study}

\begin{table}[]
\centering
\resizebox{\textwidth}{!}{%
\begin{tabular}{@{}llcccc@{}}
\toprule \toprule
\textbf{Type}                    & \textbf{Task (\# of cases)}                       & \textbf{CPM-1}          & \textbf{PLUG}        & \textbf{PanGu-$\alpha$}       & \textbf{ERNIE 3.0}      \\ \midrule
\multirow{3}{*}{Question Answering}              & Factual QA (30)        & \textbf{1.67/1.50/1.03} & 1.23/0.83/0.27       & 1.60/1.07/0.60        & \textbf{1.67/1.50/1.03} \\
                                 & Opinion QA (30)          & 1.27/0.80/-             & 1.43/1.13/-          & 1.60/1.23/-          & \textbf{1.67/1.33/-}    \\
                                 & Reasoning  (30)                          & 1.20/0.83/\textbf{0.27}          & 1.03/0.83/0.07       & 1.03/0.83/0.00       & \textbf{1.70/1.60/}0.23 \\ \midrule
\multirow{2}{*}{Interpretation}  & Interpretation of Terms (30)             & 1.23/0.73/0.70          & 1.50/0.97/0.80       & 1.57/0.97/0.70       & \textbf{1.83/1.60/1.33} \\
                                 & Reverse Dictionary (30)                  & 0.11/0.11/0.07          & 1/0.86/0.36          & 1.32/1.00/\textbf{1.00}       & \textbf{1.43/1.32}/0.93 \\ \midrule
\multirow{2}{*}{Dialogue}        & Single-Turn Dialogue (30)                & 1.63/\textbf{0.90}/-             & 1.37/0.17/-          & 1.40/0.87/- & \textbf{1.83}/0.70/-    \\
                                 & Multi-Turn Dialogue (50)                 & 1.10/0.83/-             & 0.80/0.87/-          & 1.10/1.03/-          & \textbf{1.43/1.33/-}    \\ \midrule
\multirow{5}{*}{Text Generation} & Recipe Generation (30)                   & 0.80/0.63/-             & \textbf{1.67}/1.03/- & 1.40/1.03/-          & 1.30/\textbf{1.10}/-    \\
                                 & Novel Generation (50)                    & 0.87/0.93/-             & 1.20/1.00/-          & 1.23/1.03/-          & \textbf{1.27/1.13/-}    \\
                                 & Professional Manuscripts Generation (50) & 0.97/0.83/-             & \textbf{1.37}/1.07/- & 1.23/0.83/-          & 1.33/\textbf{1.10}/-    \\
                                 & Couplet Generation (30)                  & 0.73/0.60/-             & 0.77/0.86/-          & 1.10/0.90/-          & \textbf{1.50/1.47/-}    \\
                                 & Poetry Generation (30)                   & 1.80/1.60/-             & 1.17/1.00/-          & 1.833/1.07/-         & \textbf{1.87/1.30/-}    \\ \midrule
Summarization                    & Chinese News Summarization  (30)         & 1.21/1.10/-             & 0.93/0.86/-          & 1.24/1.03/-          & \textbf{1.41/1.31/-}    \\ \midrule
\multicolumn{2}{l}{\textbf{Average}}                                   & 1.03/0.81/0.52          & 1.21/0.95/0.375      & 1.38/1.00/0.58       & \textbf{1.54/1.34/0.88} \\ \bottomrule \bottomrule
\end{tabular}%
}
\caption{The zero-shot generation performance manually evaluated on our collected 450 cases. (we reported the average score of \textit{coherence}, \textit{fluency}, and \textit{accuracy} respectively on a scale of
{[}0, 1, 2{]})}
\label{tab:zero-shot-generation}
\end{table}

%  \begin{figure*}[ht]
% 	\centering
% 	\includegraphics[width=1.0\textwidth]{examples.png}
% 	\caption{Illustrations of zero-shot generations from ERNIE 3.0.} \label{tab:gen-examples}
% \end{figure*}

% \usepackage{booktabs}
% \usepackage{graphicx}
\begin{table}[]
\centering
\resizebox{\textwidth}{!}{%
\begin{tabular}{@{}ll@{}}
\toprule \toprule
\textbf{Input (Factural QA)}                 & \begin{CJK}{UTF8}{gbsn}问题:发明加减消元法的数学家所在国家有哪些品牌汽车？ 答: \end{CJK}                                                                                                                                                    \\ \midrule
\textbf{Generation}                 & \parbox{0.8\textwidth}{\begin{CJK}{UTF8}{gbsn}有很多，比如说奔驰、宝马、奥迪、大众、保时捷。\end{CJK}}              \\ \midrule
\textbf{Input (Opinion QA)}         & \begin{CJK}{UTF8}{gbsn}问题:宝宝眼睛流泪是怎么回事？ 答:\end{CJK}                                                                                                                                                      \\ \cmidrule(l){2-2} 
\textbf{Generation}                 & \parbox{0.8\textwidth}{\begin{CJK}{UTF8}{gbsn}这种情况可能有三个原因：一是宝宝泪腺的炎症。二是宝宝的眼睛有倒睫毛或者睫毛刺激眼球引起流泪。三是宝宝有结膜炎。所以如果宝宝眼睛流泪，首先要排除宝宝倒睫毛刺激眼球引起的流泪，如果不是倒睫毛引起的流泪，还要考虑有没有泪腺炎症或者结膜炎的可能。\end{CJK}}                         \\ \midrule
\textbf{Input (Reverse Dictornary)} & \begin{CJK}{UTF8}{gbsn}问题:描写"人的品质优良"的词语有哪些？ 答:\end{CJK}                                                                                                                                                 \\ \cmidrule(l){2-2} 
\textbf{Generation}                 & \begin{CJK}{UTF8}{gbsn}品质优良,品德高尚,诚实,忠厚,老实\end{CJK}                                                                                                                                                      \\ \midrule
\textbf{Input (Novel Generation)}   & \parbox{0.8\textwidth}{\begin{CJK}{UTF8}{gbsn}江西孟龙潭与朱孝廉客都中，偶涉一兰若，殿宇禅舍，俱不甚弘敞，惟一老僧挂褡其中。见客入，肃衣出迓，导与随喜。殿中塑志公像，两壁画绘精妙，人物如生。东壁画散花天女，内一垂髫者，拈花微笑，樱唇欲动，眼波将流。\end{CJK}}                                           \\ \cmidrule(l){2-2} 
\textbf{Generation}                 & \parbox{0.8\textwidth}{\begin{CJK}{UTF8}{gbsn}西壁画一小沙弥，面目奇古，双手合十，似在祈祷。客大骇，请问之。僧曰：“此乃佛经也。”客曰：“此何所谓？”僧曰：“此有佛性，无所不有。”客曰：“何以见之？”僧曰：“佛性即是佛性。佛性无所不有，所以不能言说。”客大悟，遂去。\end{CJK}}                                  \\ \midrule
\textbf{Input (Couplet Generation)} & \begin{CJK}{UTF8}{gbsn}上联:口十心思，思乡，思友，思父母  下联:\end{CJK}                                                                                                                                                  \\ \cmidrule(l){2-2} 
\textbf{Generation}                 & \begin{CJK}{UTF8}{gbsn}眼四面观，观景，观人，观自己\end{CJK}                                                                                                                                                         \\ \midrule
\textbf{Input (News Summarization)} & \parbox{0.8\textwidth}{\begin{CJK}{UTF8}{gbsn}花盆里有黑色小飞虫，这应该是夏天大家都比较头疼的问题。灭小飞虫方法很多，给大家分享几种。电蚊拍电，粘虫板粘，杀虫剂喷，烟蒂水灌根，诱杀法，其他异味水灌根。消灭花盆里的小黑飞虫，可以把烟蒂水灌根和电蚊拍拍打结合起来，坚持一段时间就能彻底消灭。大蒜香烟泡一泡，用这一碗水，小虫去无踪。文章标题是《\end{CJK}} \\ \cmidrule(l){2-2} 
\textbf{Generation}                 & \parbox{0.8\textwidth}{\begin{CJK}{UTF8}{gbsn}花盆里有黑色小飞虫怎么办？》\end{CJK}}                                                                     \\ \bottomrule \bottomrule
\end{tabular}%
}
\caption{Illustrations of zero-shot generations from ERNIE 3.0.}
\label{tab:gen-examples}
\end{table}

We manually collected 450 cases to evaluate the zero-shot generation ability of current large-scale pre-trained models on 13 tasks from 5 different types including Question Answering, Interpretation, Dialogue, Text Generation and Summarization. In human evaluation, the annotators are asked to score the generation quality on a scale of $[0, 1, 2]$. We reported the average score of \textit{coherence}, \textit{fluency}, and \textit{accuracy} in Tab.~\ref{tab:zero-shot-generation}, and showed some zero-shot generations of ERNIE 3.0 in Tab.~\ref{tab:gen-examples}. ERNIE 3.0 can generate the most coherent, fluent and accurate texts on average as compared to CPM-1, PLUG, PanGu-$\alpha$~\footnote{We use the implementation of CPM-1 in \url{https://github.com/jm12138/CPM-Generate-Paddle}, PLUG in \url{https://nlp.aliyun.com/portal?/BigText_chinese\#/BigText_chinese} and PanGu-$\alpha$ in \url{https://git.openi.org.cn/PCL-Platform.Intelligence/PanGu-Alpha}}. The introduction of three scoring metrics are listed as follows, and the scoring details are provided in Tab.~\ref{tab:score-detail}.
\begin{itemize}
    \item \textbf{Coherence} measures whether the generation is relevant and consistent with the context.
    \item \textbf{Fluency} evaluates whether the generated text is natural or readable. A fluent text should have no semantic contradiction among the generated text.
    \item \textbf{Accuracy} is a metric to evaluate whether the generated text is the same as the ground truth.
\end{itemize}

\begin{table}[]
\centering
\resizebox{\textwidth}{!}{%
\begin{tabular}{@{}c|lll@{}}
\toprule \toprule
\textbf{Score} & \textbf{Coherence}                                                                                                                                    & \textbf{Fluency}                                                                                                                                 & \textbf{Accuracy}             \\ \midrule
0              & \begin{tabular}[c]{@{}l@{}}The generation is not related to the context.\\ The generation has obvious conflicts with the context.\end{tabular}        & \begin{tabular}[c]{@{}l@{}}The generation is unnatural.\\ There are contradictions in the generated text.\end{tabular}                           & The answer is wrong.          \\ \midrule
1              & \begin{tabular}[c]{@{}l@{}}The generation is weakly related to the context.\\ The generation has minor logic conflicts with the context.\end{tabular} & \begin{tabular}[c]{@{}l@{}}The generation has minor influent part.\\ The generation slightly influences the reading.\end{tabular}                & The answer is partly correct. \\ \midrule
2              & \begin{tabular}[c]{@{}l@{}}The generation is strongly related to the context.\\ The logic in the generation is aligned with the context.\end{tabular} & \begin{tabular}[c]{@{}l@{}}The generation is semantically complete and fluent.\\ There are no contradictions in the generated text.\end{tabular} & The answer is correct.        \\ \bottomrule \bottomrule
\end{tabular}%
}
\caption{Scoring details for zero-shot generation.}
\label{tab:score-detail}
\end{table}

\subsection{Experiments on SuperGLUE}

	\begin{table*}[h]
		\begin{center}
			\scalebox{1.0}{
				\begin{tabular}{
						l|cccccccc|c
					}
				\toprule \toprule
            \textbf{Model}          & \textbf{BoolQ} & \textbf{CB}        & \textbf{COPA} & \textbf{MultiRC}   & \textbf{ReCoRD}    &\textbf{RTE}  & \textbf{WiC}  & \textbf{WSC} & \textbf{Score}  \\
				\midrule
            Human Baseline  & 89.0  & 95.8/98.9 & 100  & 81.8/51.9 & 91.7/91.3 & 93.6 & 80.0 & 100  & 89.8 \\
				\midrule \midrule
            T5+Menna        & \textbf{91.4}  & 95.8/97.6 & 98.0 & 88.3/63.0 & 94.2/93.5 & 93.0 & \textbf{77.9} & 96.6 & 90.4 \\
            DeBERTa        & 90.4  & 95.7/97.6 & \textbf{98.4} & 88.2/\textbf{63.7} & 94.5/94.1 & \textbf{93.2} & 77.5 & 95.9 & 90.3   \\
				\midrule
            ERNIE 3.0       & 91.0  & \textbf{98.6/99.2} & 97.4 & \textbf{88.6}/63.2 & \textbf{94.7/94.2} & 92.6 & 77.4 & \textbf{97.3} & \textbf{90.6} \\
            
					\bottomrule \bottomrule
				\end{tabular}
			}
			\caption{SuperGLUE test set results which are scored by the SuperGLUE evaluation server (Results are recorded at July 3, 2021 from https://super.gluebenchmark.com/leaderboard).
			}
		    \label{tab:superglue}
		\end{center}
		\vspace{-5mm}
	\end{table*}

As a multi-task benchmark for natural language understanding, SuperGLUE~\cite{wang2019superglue} is usually used to evaluate the performance of pre-training models. We also test the performance of ERNIE 3.0 on SuperGLUE, which covers a diverse range of NLP datasets as follows.

\begin{itemize}

\item BoolQ (Boolean Questions, \cite{chen2018bq}) is a QA task where each example consists of a short passage and a yes/no question about the passage. The task is to answer the questions with YES or NO, and the metric of this task is accuracy.

\item CB (Commitment Bank, \cite{demarneffe:cb}) is an imbalanced corpus of natural language inference task. The task is evaluated using accuracy and macro-F1.
 
\item COPA (Choice of Plausible Alternatives \cite{roemmele2011choice}) is a causal reasoning task based on common sense knowledge. The data are curated from blogs and a photography-related encyclopedia. Following the original work, we evaluate this task using accuracy.

\item  MultiRC (Multi-Sentence Reading Comprehension \cite{khashabi2018looking}) is a QA task where each example consists of a context paragraph, a question about that paragraph, and a list of possible answers. The system must predict which answers are true and which are false. The evaluation metrics are F1 over all answer-options (F$1_{a}$) and exact match of each question’s set of answers (EM).

\item ReCoRD (Reading Comprehension with Commonsense Reasoning Dataset, \cite{zhang2018record}) is a multiple-choice QA task. It requires the model to pick an entity to complete the answer, given a context of news article and a Cloze-style question. This task is evaluated with max (over all mentions) token-level F1 and exact match.
  
\item RTE (Recognizing Textual Entailment \cite{dagan2006pascal}) dataset comes from a series of annual competitions on textual entailment. It is a natural language inference corpus and evaluated with accuracy.

\item WiC (Word-in-Context \cite{pilehvar2018wic}) is a word sense disambiguation task cast as binary classification of sentence pairs using accuracy as the evaluation metrics.

\item WSC (Winograd Schema Challenge \cite{levesque2011winograd}) is a coreference resolution task in which examples consist of a sentence with a pronoun and a list of noun phrases from the sentence as choices. The system must select the correct referent of the pronoun from the provided choices. This task is evaluated with accuracy.

\end{itemize}

Similar to the pre-training corpus used in RoBERTa~\cite{liu2019roberta} and DeBERTa~\cite{he2020deberta}, we compiled the English pre-training corpus for ERNIE 3.0 including English Wikipedia, BookCorpus~\cite{zhu2015aligning}, CC-News~\cite{ccnews}, OpenWebText~\cite{openwebtext}, Stories~\cite{trinh2018simple}. As shown in the Table \ref{tab:superglue}, ERNIE 3.0 surpasses T5~\cite{T5} and DeBERTa~\cite{he2020deberta} and obtains a score of 90.6, taking the first place in SuperGLUE Benchmark.

\section{Analysis}\label{sec:analysis}
\begin{wrapfigure}{r}{0.4\textwidth}
    \begin{center}
    \includegraphics[width=0.4\textwidth]{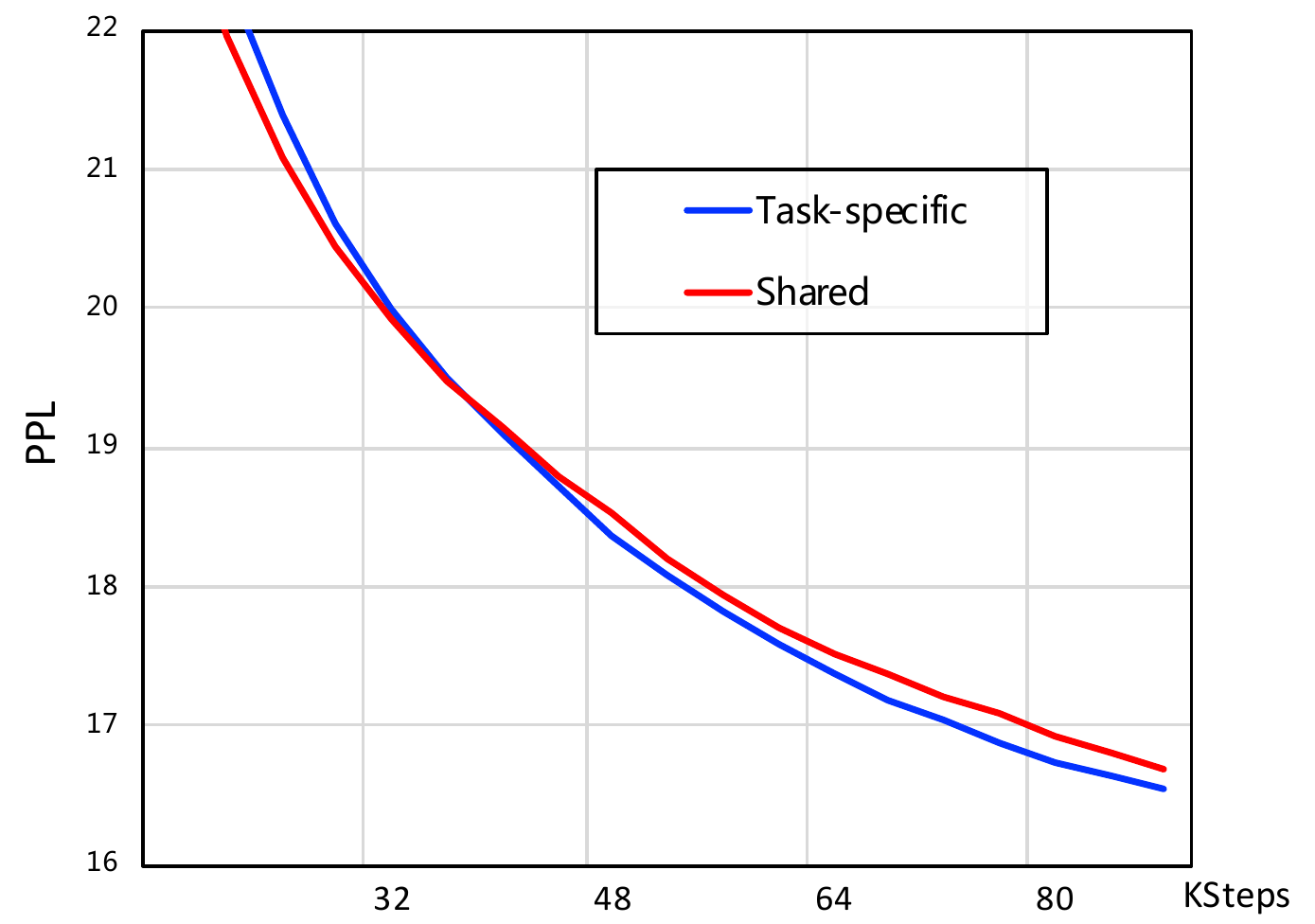}
    \end{center}
    \caption{Perplexity variation of the NLG pre-training task with respect to training steps.}
    \label{fig:ablation-branch}
\end{wrapfigure}
\paragraph{The Effectiveness of the Task-specific Representation Modules}
To verify the effectiveness of the task-specific networks, we compare our proposed structure with those which share parameters under various pre-training tasks. For the ablation test, we choose understanding and generation as two different training paradigms and utilize the corresponding tasks mentioned in Section \ref{sec:pre-training tasks}. The unified network follows the base model settings (12 layers, 768 dims, 12 attention heads), and the task-specific networks for each task paradigms are set to 3 layers, 256 dims, and 4 attention heads. For the contrast model, the task-specific network is shared across different task paradigms. Figure \ref{fig:ablation-branch} illustrates the perplexity variation of the NLG task during the pre-training process. 

As shown in Figure \ref{fig:ablation-branch}, the model with its own task-specific network for different task paradigms reaches a higher convergence speed. Furthermore, as training progresses, the performance gap becomes bigger compared to the model with a shared task-specific network. The experimental result shows the effectiveness of the proposed task-specific networks and demonstrates the necessity of distinguishing different tasks.

\paragraph{Universal Knowledge-Text Prediction}
A group of ablation experiments is conducted to evaluate the performance of the universal knowledge-text prediction task. The relation extraction task is a typical knowledge-driven task, aiming to predict the relationship between two entities mentioned in a given sentence. Specifically, we add four special tokens, \textsc{[HD]}, \textsc{[/HD]}, \textsc{[TL]} and \textsc{[/TL]} to identify the mention of a head entity and a tail entity respectively, then the relation classification is performed on the sum of the final representations of the aforementioned four special tokens. We construct the experiments on SanWen and FinRE datasets and as shown in Table \ref{tab:uktp}, the knowledge enhancement strategy achieves impressive empirical performance on the relation extraction task. 

\begin{wraptable}{r}{0.45\textwidth}
    \centering
	\begin{tabular}{lll}
		\toprule \toprule
		\textbf{Dataset}     & \textbf{ERNIE$_\text{Base}$}    & \textbf{ERNIE$_\text{Base}$+UKTP}  \\ 
		\midrule
		SanWen  & 75.56 &77.36(+1.80)     \\ 
		FinRE   & 58.19 &59.75(+1.56)   \\ 
		\bottomrule \bottomrule
	\end{tabular}
	\caption{Ablation experiments of universal knowledge-text prediction task.}
	\label{tab:uktp}
\end{wraptable}

%the final representations of four special tokens are summed to preform relation classification. 

In addition, the zero-shot generation experiment on CKBQA also confirms the effectiveness of the universal knowledge-text prediction task. Specifically, the knowledge-based question answering (KBQA) task requires a model to search and reason for correct answers based on a knowledge graph. It's suitable to measure the knowledge learning capability of the pre-trained languages models using the KBQA task. We use the "QUESTION: \$QUESTION?  ANSWER:" as the prompt for zero-shot learning and then compare the performance of our proposed model with several state-of-the-art pre-trained language models on the CKBQA dataset. As shown in Table \ref{tab:zero-shot-result}, ERNIE 3.0 significantly outperforms PanGu-$\alpha$ and CPM-1 in the CKBQA dataset which indicates that ERNIE 3.0 has the ability to memorize and learn more knowledge.

\paragraph{Progressive Learning to Speed up Convergence}

 We record the training convergence speed on two architecture settings including $\mathrm{ERNIE_{Base}}$ and $\mathrm{ERNIE_{1.5B}}$, in which the architecture settings of $\mathrm{ERNIE_{Base}}$ follows \cite{sun2019ernie} and
 \begin{wraptable}{r}{0.45\textwidth}
	\centering
	\begin{tabular}{ll}
		\toprule \toprule
		\textbf{Method}         & \textbf{Training Time} \\ 
		\midrule
		ERNIE$_\text{Base}$   & 11h30m    \\ 
		+Progressive Learning    & 4h(-65.21\%)    \\ 

		\midrule
		ERNIE$_\text{1.5B}$          & 5h55m   \\ 
		+Progressive Learning   & 3h4m(-48.2\%)   \\ 
		\bottomrule \bottomrule
	\end{tabular}
	\caption{Progressive Learning To Speedup Training.}
	\label{tab:4dpro}
\end{wraptable}
 $\mathrm{ERNIE_{1.5B}}$ model consists of 48 layers with a hidden size of 1,536 and 24 attention heads. As shown in Tab.~\ref{tab:4dpro}, we record the time for the loss value of the model converges to the same as that of the ERNIE 3.0. For the $\mathrm{ERNIE_{Base}}$ model, the convergence time is reduced by 65.21\% from 11 hours to 4 hours, and for the $\mathrm{ERNIE_{1.5B}}$, the convergence time is reduced by 48\%. For both two settings, we carry out pre-training on 8 NVIDIA Tesla V100 GPUs. For $\mathrm{ERNIE_{Base}}$, we increased the batch size from 8 to 2048 and the sequence length from 128 to 512, the learning rate increases linearly from 0 to 1e-4, and the dropout keeps 0 in the progressive warmup stage. For $\mathrm{ERNIE_{1.5B}}$, we gradually increase the batch size from 8 to 8192, The learning rate increases from 0 to 6e-4, the dropout also keeps 0. The rest settings for the experiment remain as same as \cite{sun2019ernie}. For $\mathrm{ERNIE_{1.5B}}$, to achieve the peak batch size within the constraint of GPU memory, the gradient accumulation strategy is used during the pre-training. 

\section{Conclusion}
%We present ERNIE 3.0, a knowledge enhanced 10-billion parameter pre-trained model trained on 4T text data and a knowledge graph containing 50 millions facts.
we proposed the ERNIE 3.0 framework to pre-train a knowledge enhanced 10-billion parameter model on a 4TB corpus including plain texts and a knowledge graph. In order to handle both language understanding and generation tasks with zero-shot learning, few-shot learning and fine-tuning, ERNIE 3.0 designs a unified pre-training framework that integrates both auto-encoder networks and auto-regressive networks. We construct extensive experiments on various datasets from different task paradigms and fields, and the results demonstrate the effectiveness of ERNIE 3.0 as compared to the previous state-of-the-art pre-trained models.

\bibliographystyle{unsrt}  
\bibliography{references}  

\newpage

\end{document}